%% file: 2018-CVIU-FeatureImportance.tex
\documentclass[times,twocolumn,final,authoryear]{elsarticle}

\usepackage{ycviu}
\usepackage{framed,multirow}

\usepackage{amssymb}
\usepackage{latexsym}

\usepackage{url}

\usepackage[utf8]{inputenc}
\usepackage{times}
\usepackage{epsfig}
\usepackage{booktabs}
\usepackage{amssymb}
\usepackage{amsfonts}
\usepackage{enumitem}
\usepackage{wrapfig}
\usepackage[table,usenames,dvipsnames,svgnames]{xcolor} 
\usepackage[breaklinks=true,bookmarks=false]{hyperref}
\hypersetup{
	colorlinks=true,
	linkcolor=BrickRed,
	citecolor=OliveGreen,
	filecolor=magenta,
	urlcolor=cyan
}

\definecolor{newcolor}{rgb}{.8,.349,.1}

\graphicspath{{./images/}}

\usepackage[utf8]{inputenc}
\usepackage{amsmath}

\usepackage{blindtext}

\newcommand{\tmpframe}[1]{\fbox{#1}}
\renewcommand{\tmpframe}[1]{#1}

\newcommand{\first}[1]{\textcolor{BrickRed}{\textbf{#1}}}
\newcommand{\second}[1]{\textcolor{ForestGreen}{\textbf{#1}}}
\newcommand{\third}[1]{\textcolor{blue}{\textbf{#1}}}

\newcommand{\new}[1]{\textcolor{black}{#1}}

\newcommand{\dataset}{CarsReId74k}
\newcommand{\vehiclesAll}{17,681}
\newcommand{\tracksAll}{73,976}
\newcommand{\pairsAll}{277,236}
\newcommand{\camerasAll}{66}
\newcommand{\negativesAll}{1283}
\newcommand{\imagesAll}{3,242,713}

\newcommand{\avgFeatCars}{1536}

\newcommand{\SecondRev}[1]{\textcolor{blue}{[\textbf{R2:} #1]}}

\journal{Computer Vision and Image Understanding}

\begin{document}

\begin{frontmatter}

\title{Learning Feature Aggregation in Temporal Domain for Re-Identification}

\author[1]{Jakub \snm{Špaňhel}\corref{cor1}}
\cortext[cor1]{Corresponding author: 
	Tel.: +420-54114-1298;}
\ead{ispanhel@fit.vutbr.cz}

\author[1]{Jakub \snm{Sochor}\fnref{fn1}}
\fntext[fn1]{Joint first author}


\author[1]{Roman \snm{Juránek}}
\author[1]{Petr \snm{Dobeš}}
\author[1]{Vojtěch \snm{Bartl}}
\author[1]{Adam \snm{Herout}}

\address[1]{Graph@FIT, Brno University of Technology, Faculty of Information Technology, Centre of Excellence IT4Innovations }

\received{1 May 2013}
\finalform{10 May 2013}
\accepted{13 May 2013}
\availableonline{15 May 2013}
\communicated{S. Sarkar}

\input{2018-CVIU-FeatureImportance-0-Abstract}


\end{frontmatter}



\input{2018-CVIU-FeatureImportance-1-Introduction}
\input{2018-CVIU-FeatureImportance-2-SOTA}

\input{2018-CVIU-FeatureImportance-3-Methodology}
\input{2018-CVIU-FeatureImportance-4-Dataset}
\input{2018-CVIU-FeatureImportance-5-Results}

\input{2018-CVIU-FeatureImportance-6-Conclusions}

\section*{Acknowledgments}
\noindent
This work was supported by TACR project ``SMARTCarPark'', TH03010529. Also, this work was supported by The Ministry of Education, Youth and Sports of the Czech Republic from the National Programme of Sustainability (NPU II); project IT4Innovations excellence in science - LQ1602.

\section*{Notice}
\noindent
This paper is under consideration at Computer Vision and Image Understanding

\bibliographystyle{model2-names}
\bibliography{2018-CVIU-FeatureImportance-bibliography}

\end{document}

%% file: 2018-CVIU-FeatureImportance-0-Abstract.tex
\begin{abstract}
	Person re-identification is a standard and established problem in the computer vision community. In recent years, vehicle re-identification is also getting more attention. 
	In this paper, we focus on both these tasks and propose a~method for aggregation of features in temporal domain as it is common to have multiple observations of the same object. The aggregation is based on weighting different elements of the feature vectors by different weights and it is trained in an end-to-end manner by a Siamese network. The experimental results show that our method outperforms other existing methods for feature aggregation in temporal domain on both vehicle and person re-identification tasks.
	Furthermore, to push research in vehicle re-identification further, we introduce a novel dataset \dataset. The dataset is not limited to frontal/rear viewpoints. It contains \vehiclesAll\ unique vehicles, \tracksAll\ observed tracks, and \pairsAll\ positive pairs. The dataset was captured by \camerasAll\ cameras from various angles. 
\end{abstract}

%% file: 2018-CVIU-FeatureImportance-1-Introduction.tex
\section{Introduction}

We consider the problem of re-identification of individuals observed by different ca\-me\-ras at different locations and times.  Our work applies to the fairly standard person re-identification \citep{Wang2014,Hirzer2011,Xu2017,Zhang2017,Chen2017,Zhou2017}, and to the rather emerging vehicle re-id \citep{Liu2016CVPR,Liu2016deep,Shen2017,Wang2017,Yan2017,Zhang2017improving}, but it can be used for other similar tasks as well.

The re-id system is given a query track of images and a~database of pre-stored tracks, one of which is assumed to share the same identity with the query.  The system is supposed to output a small subset of the best matching database samples along with their similarity scores.
Some solutions process the images in the tracks directly (comparing images in the query track versus images in the database -- e.g. \cite{Zapletal2016vehicle}). However, fast and real-time processing requires the system to extract a short feature vector for each of the database tracks and to match them to the feature vector extracted from the query track by computing a cheap pairwise metric.  Our work is targeted on the second, generally more efficient, mode of processing, \new{that is extraction of a single fixed-size feature vector for a track of variable length by aggregating the feature vectors extracted from individual observations (images).}

\begin{figure}
	\centering
	\includegraphics[width=0.95\linewidth]{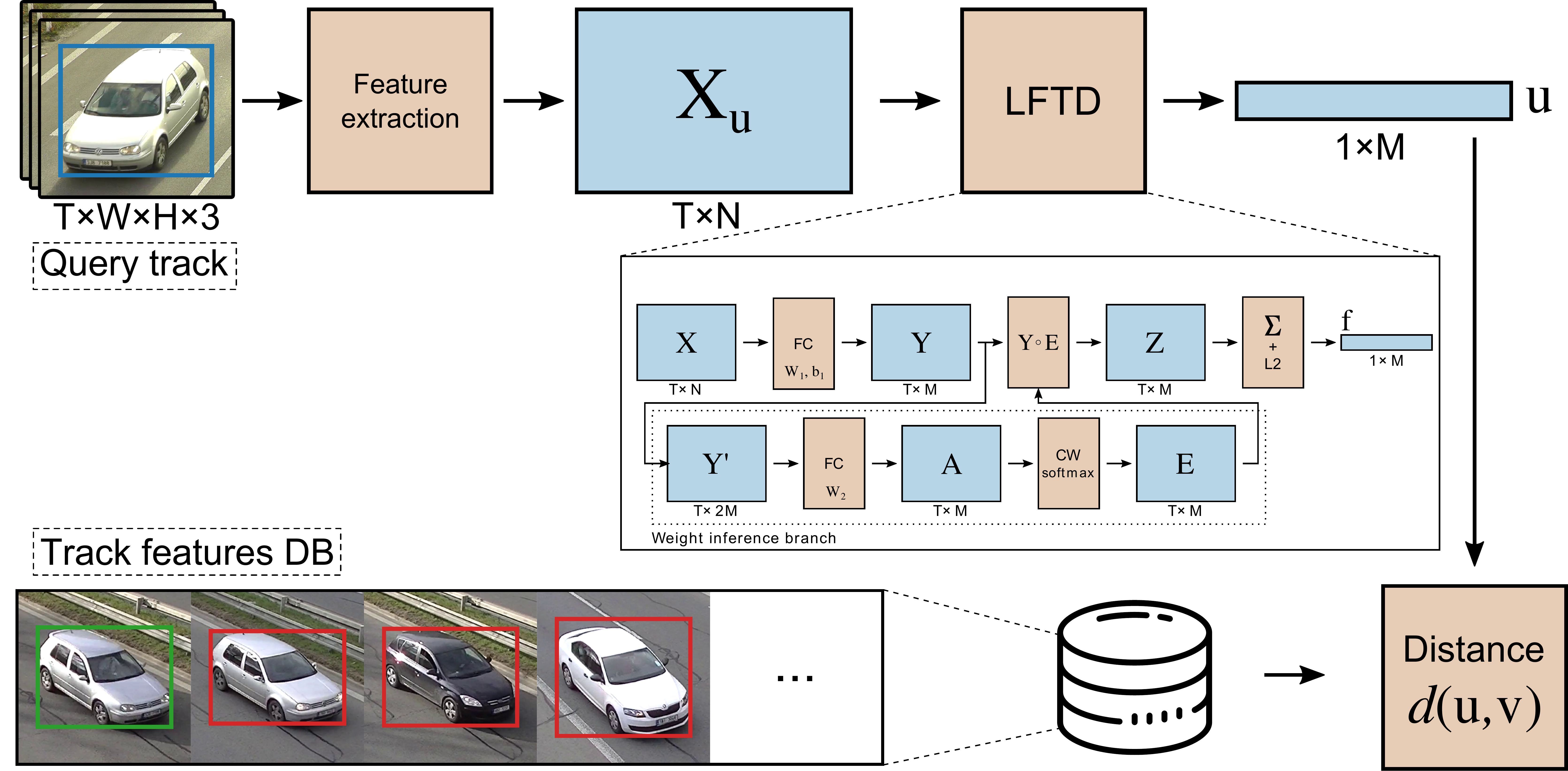}
	\caption{We propose a new method \textbf{LFTD} for aggregation of features in temporal domain. The method generates one feature vector per track of observed objects (e.g. vehicles, persons). See Section \ref{sec:MethodologyFeatureAggregation} for details.}
	\label{fig:Teaser}
\end{figure}

We propose a new method for \textbf{feature aggregation in temporal domain} LFTD (Learning Features in Temporal Domain) which takes feature vectors extracted from the individual observations (images) as its input, and it results in a single relatively low-dimensional time-pooled feature vector usable by the re-id system.
Unlike other methods which use either RNN \citep{McLaughlin2016,Zhang2017,Yan2016,Chen2017,Xu2017,Zhou2017,Zhang2017learning} or produce weights for feature vectors as a~whole \citep{Yang2017,Zhou2017,Xu2017}, our method produces a different weight for every element of the feature vectors which leads to an improved performance as different parts of feature vectors are weighted differently. The weights are generated by a neural network for each set (track) of feature vectors. The final feature vector for the track is obtained by computing element-wise product between the track's features and the weight matrix, and then reducing the matrix in temporal domain by summation. The results show that the proposed method outperforms other methods \citep{Yan2016,McLaughlin2016,Gao2016,Xu2017,Zhang2017,Chen2017,Zhou2017,Zhang2017learning} in both vehicle and person re-identification tasks. See Figure~\ref{fig:Teaser} for the full re-id pipeline.

Furthermore, we propose to use a different metric for comparing the feature vectors.
Previous works \citep{Kostinger2012,Liao2015,Shi2016} showed that it is beneficial to use Mahalanobis distance for feature comparison rather than Euclidean (or cosine) distance. However, the Mahalanobis distance has significant limitations, mainly its time complexity which is quadratic with respect to feature vector dimensionality.
Therefore, we propose to use \textbf{Weighted Euclidean} distance, constructed by constraining the Mahalanobis distance learning to diagonal matrix.  The experiments show that it outperforms both Mahalanobis \citep{Shi2016} and Euclidean distance, while it keeps linear time and memory complexity.

To improve the availability of datasets for vehicle re-identification, we collected and annotated a new vehicle re-identification dataset called \textbf{\dataset}. As it is common in traffic surveillance to have whole tracks of vehicles and not individual images, the dataset includes multiple observations for each vehicle as it is passing in front of the cameras (\emph{left}, \emph{center}, \emph{right}).
We focus on \textbf{appearance-based} vehicle re-identification: vehicles' license plates were only used for ground truth data acquisition (recorded by a \emph{zoomed-in} camera).
The images of vehicles taken by the other cameras are in most cases so small that it is not possible to recognize the license plates. 
%
The dataset contains \vehiclesAll\ unique vehicles, \tracksAll\ observed tracks, and \pairsAll\ positive pairs, taken by \camerasAll\ cameras from various angles in multiple sessions. We make the dataset publicly available\footnote{\url{https://medusa.fit.vutbr.cz/traffic}} for future comparison and research.


%% file: 2018-CVIU-FeatureImportance-2-SOTA.tex
\section{Related Work}

\subsection{Image Feature Pooling in Temporal Domain}
In this section, we provide an overview of existing methods for feature pooling (aggregation) in temporal domain. Such pooling is usually used in the context of person re-identification (with the exception of \cite{Yang2017} who used it for video face recognition). The methods are often trained by using a Siamese network \citep{McLaughlin2016,Zhang2017,Yan2016,Chen2017,Xu2017,Yang2017} with contrastive loss and optionally identification loss as well.

\cite{McLaughlin2016} propose an approach for temporal domain pooling based on Recurrent Neural Networks (RNN). The authors extract features using a~CNN and use a~recurrent layer to compute the features for the whole track. The used RNN has an output for each time step and these outputs are averaged to obtain the final feature vector for re-identification. The authors further propose to  use optical flow as an additional input to the network.  A similar approach was proposed by \cite{Zhang2017} with the exception that their method uses bi-directional RNN to get better re-id results.  Also, the method proposed by \cite{Yan2016} is similar with the exception that the image level features are not trained and LBP and color features are used instead.

\cite{Chen2017} also follow the work of \cite{McLaughlin2016}. However, they propose to merge the features extracted by RNN together with CNN spatial features averaged over the time steps.
The authors use three such networks for different body parts and fuse their output features (by a weighted sum).

Another approach based on the work by \cite{McLaughlin2016} is proposed by \cite{Xu2017}, who introduce significant modifications to the method. First, image level features are extracted by spatial pyramid pooling; thus, spatial information is preserved in the feature vector. These features are then fed into a recurrent layer (similar to \cite{McLaughlin2016}). Finally, the recurrent features are pooled by an Attentive Temporal Pooling layer proposed by the authors.
However, a significant drawback of the proposed method is that it requires both the query and the gallery raw feature vector sequences during distance computation, leading to more complex processing during the search in the database.

\new{\cite{Zhang2017learning} adds a feature pooling layer into the CNN architecture before the first fully connected layer. This layer aggregates key information from different views of the person's trajectory (different time steps) in a single feature vector. They also incorporate two different learning distance metrics -- minimum distance and average of minimum distance for comparing the query track with tracks in the database.}

Unlike the other authors, \cite{Yang2017} focus on video face recognition. The authors propose an approach to temporal pooling based on weighting of feature vectors from different time steps. The weight for a feature vector is obtained as a dot product with a template, which is computed by a fully connected layer. The weights are then normalized to a probability distribution by softmax function. The weights for different time steps scale the contributions of images in the sequence according to their dis\-cri\-mi\-na\-tive value.

Similarly, \cite{Zhou2017} propose to use a temporal attention model and ge\-ne\-rate weights for feature vectors in the track. However, in contrast to \cite{Yang2017}, the weights are generated at each time step for all the feature vectors in the sequence. Then, at all time steps, all feature vectors (from the given sequence) are weighted by a~set of (different) weights; thus, at each time step, differently weighted input feature vectors are produced. The weights at each time step are obtained by a RNN layer. Furthermore, similarly to \cite{McLaughlin2016}, the weighted feature vectors are fed into another RNN layer with output at each time step and then averaged to obtain the final track re\-pre\-sen\-ta\-tion. The authors also use spatial RNNs to further improve the re-identification results.


Generally, for temporal pooling, the authors use either recurrent neural networks \citep{McLaughlin2016,Zhang2017,Yan2016,Chen2017,Xu2017}, learned weighting of feature vectors \citep{Yang2017}, or a combination of these app\-roa\-ches \citep{Zhou2017}. In contrast to the described methods, the proposed method produces a different weight for every element of the feature vectors.

\subsection{Person Re-Identification}

Besides standard deep features learned by a Siamese network \citep{McLaughlin2016,Zhang2017,Yan2016,Chen2017,Zhou2017}, other approaches to person re-identification have been proposed.

Several papers proposed to use body parts \citep{Cheng2016,Khan2017,Li2017,Zhao2017}. Other papers went beyond Siamese networks and proposed triplet loss \citep{Cheng2016,Hermans2017} or quadruplet loss \citep{Chen2017Quadruplet}. There were also attempts to learn a metric for the re-identification like KISSME~\citep{Kostinger2012}, XQDA \citep{Liao2015}, \cite{You2016} learn Mahalanobis distance on LBP and HOG3D features, and finally \cite{Shi2016} learn Mahalanobis distance in an end-to-end manner. \cite{Sun2017svdnet} proposed to use SVD for weight matrix orthogonalization to de-correlate feature vectors for person re-id.

Other authors exploit different types of features. For example, \cite{Wu2016} propose to use deep features learned by a CNN together with hand-crafted features. The final representation for an image is obtained by fusing these features. \cite{Matsukawa2016} use a novel descriptor based on hierarchical gaussians computed for patches in image. \cite{Chen2017Fast} propose to use compact binary hash codes as features for fast person re-identification. There were also attempts \citep{Liu2015,Gao2016} to recognize the walking cycle in image sequence and use the walking cycle to improve the accuracy of re-identification.

A group of works also propose to replace different parts of the re-identification pipeline by alternative solutions. \cite{Zhong2017} use re-ranking based on $k$-reciprocal nearest neighbors to improve the performance. \cite{Zhou2017Point2Set} propose to use point-to-set distance instead of standard point-to-point. \cite{Lin2017} take inter-camera consistencies of id assignment into account during training and inference to boost the results of re-identification. \cite{Xiao2016} propose to use domain guided dropout to improve re-identification performance when trained on multiple datasets. \cite{Wang2016} propose to add a network computing a cross-image representation for pairs of images. \cite{Cho2016} estimate persons' poses and compare images with each person in an as similar as possible pose. \cite{Su2016} use attributes (e.g ``long sleeve'') for person re-id. The attributes are first learned on a different dataset with attributes present and then fine-tuned for the target dataset. The attributes supervision for the target dataset comes from the assumption that same person has the same (unknown) attributes.

\subsection{Vehicle Re-Identification}

There are mainly two types of methods -- methods based on automatic license plate recognition \citep{Du2013automatic,Kluwak2016,Wen2011algorithm}, which are not anonymous and require zoomed-in cameras. The other type of methods is based on vehicles' visual appearance \citep{Arth2007object,Feris2012large,Zapletal2016vehicle,Liu2016large} or on a combination of both approaches \citep{Liu2016deep}.


Formerly, different types of \emph{hand-crafted} features  were used. For example authors used PCA-SIFT \citep{Arth2007object}, HOG descriptors and color histograms \citep{Zapletal2016vehicle}, SIFT-BOW and Color Names model \citep{Liu2016large} or just information about date, time, color, speed and vehicles' dimensions \citep{Feris2012large}. Recently, \emph{deep} features learned by CNNs \citep{Liu2016CVPR,Shen2017,Wang2017,Yan2017,Zhang2017improving} were used for this task. \cite{Liu2016deep} combine the hand-crafted and deep features.





Improvements were also made by exploiting \emph{spatio-temporal} \citep{Liu2016deep,Wang2017} or \emph{visual-spatio-temporal} \citep{Shen2017} properties. Some of them benefit from Siamese CNNs for license plate verification \citep{Liu2016deep} or vehicle image similarities \citep{Shen2017}. Moreover, introduction of triplet loss \citep{Zhang2017improving} or Coupled Cluster Loss (CCL) \citep{Liu2016CVPR} led to accuracy improvements and faster convergence. Recently, \cite{Yan2017} propose to use Generalized Pairwise Ranking or Multi-Grain based List Ranking for retrieval of similar vehicles, which performs even better than CCL.

\subsection{Vehicle Re-Identification Datasets}

There are datasets of vehicles \citep{Krause2013,Yang2015,Sochor2017BoxCars}, which are created for fine-grained recognition with annotations on several attributes such as type, make and color.
However, the identities of the vehicles in the datasets are not known; thus, the datasets are not directly applicable for vehicle re-identification, especially for evaluation.

When it comes to genuine vehicle re-identification, \cite{Liu2016deep} constructed a rather small VeRi-776 dataset containing 50,000 images of 776 vehicles. \cite{Liu2016CVPR} collected VehicleID dataset containing 26,267 vehicles in 220k images taken from a frontal/rear viewpoint above road.
Recently, \cite{Yan2017} published two datasets VD1 and VD2 for vehicle re-identification and fine-grained classification with over 220k of vehicles in total, with make, model, and year annotation. However, both datasets are limited to frontal viewpoints only.

%% file: 2018-CVIU-FeatureImportance-3-Methodology.tex
\section{Proposed Method for Learning Feature Aggregation in Temporal Domain}
The standard baseline to aggregating features from multiple observations of the same object in temporal domain is to use averaging over time. However, existing literature \citep{Yan2016,McLaughlin2016,Gao2016,Xu2017,Zhang2017,Chen2017,Zhou2017} shows that the accuracy can be improved over the simple averaging by feature vector weighting or by using RNN.
We propose a novel method for the aggregation in temporal domain, which is based on weighting different elements of the features vectors by different weights.

\new{The proposed LFTD method aggregates arbitrary features from a sequence of images (of an arbitrary length), extracted by any feature extractor (it can be even some of newly presented spatial attention networks \citep{Wang2017residual,Su2017pose}) into a single fixed-sized feature vector. It allows to create a database of previously seen objects (with multiple observations) with such fixed-sized feature vectors and then quickly search the database for objects similar to query objects.
LFTD expands the feature dimensions by concatenating the average feature vector to features extracted in every time step. It allows the network to propagate global information form the track to each individual observation. Feature vectors are weighted by column-wise softmax (i.e. along time axis) which forces the network to pick important observation for every feature in the vector instead of weighting observations as a whole.
This network design performed the best during our preliminary experiments, compared with user-based vector normalization (subtracting or dividing features by average feature vector), or different types of feature expansion (e.g. by max-pooled feature vector, etc.).
}

\new{The method is detailed in the following sections.}

\subsection{Image Feature Extraction}
We are processing \emph{the whole tracks} of objects of interest with labels corresponding to identities $\left\lbrace (\mathcal{T}_i, l_i) \right\rbrace$, where $\mathcal{T}_i$ is a sequence of images $(\mathbf{I}_1, \mathbf{I}_2, \dots, \mathbf{I}_{T_i})$, i.e. observations of an object $l_i$ in the track.

For each \new{track (image sequence)}, features are extracted for each image independently by a feature extractor (a CNN-based or another one, the method is not limited by design to a particular type).
The feature extractor yields a feature matrix $\mathbf{X}_i \in \mathbb{R}^{T_i \times N}$ for each track $\mathcal{T}_i$. \new{$T_i$ is number of time samples (images) for each track $\mathcal{T}_i$} and $N$ is the length of an individual feature vector. In our experiments $N = 2048$, in case of ResNet50, and $N = 1536$ for Inception-ResNet-v2.

To make the notation uncluttered, we will omit the lower index $i$ from now on. Therefore, we will refer to a individual track as $\mathcal{T}$, the number of time samples of the track as $T$, and its features as $\mathbf{X}  \in \mathbb{R}^{T \times N}$.


\subsection{Processing of Features in Temporal Domain}
\label{sec:MethodologyFeatureAggregation}
\begin{figure}[t]
	\centering
	\includegraphics[width=\linewidth]{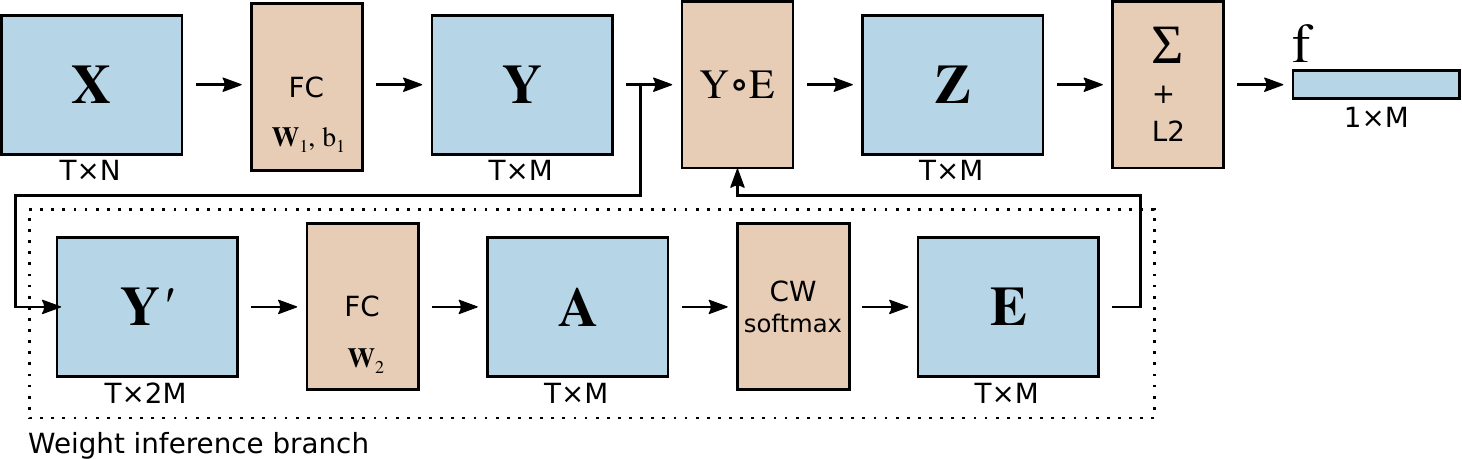}%
	\caption{Schematic network design representing the proposed method for feature aggregation in temporal domain. See Section~\ref{sec:MethodologyFeatureAggregation} for explanation of the symbols. }
	\label{fig:TrainingNetwork}
\end{figure}

The schematic design of the feature aggregation network is illustrated in Figure~\ref{fig:TrainingNetwork} and the description follows.
Aggregation of features $\mathbf{X} \in \mathbb{R}^{T\times N}$ in temporal domain is essentially a mapping $\varphi: \mathbb{R}^{T\times N} \mapsto \mathbb{R}^M$,
where $M$ is the dimensionality of feature vector $\mathbf{f}$ representing track $\mathcal{T}$.

First, the feature vector of each observation in the track is compressed from $N$ to $M$ dimensions ($M < N$) by
\begin{equation}
\mathbf{y}_\tau = \mathrm{tanh}\left(\mathbf{W}_1 \mathbf{x}_\tau + \mathbf{b}_1\right), \quad 1 \leq \tau \leq T, \label{eq:projection}
\end{equation}
where $\mathbf{W}_1 \in \mathbb{R}^{M \times N}$ are the parameters of the first fully connected layer (Figure~\ref{fig:TrainingNetwork}), forming a compressed feature matrix $\mathbf{Y} \in \mathbb{R}^{T \times M}$.

In order to allow ``communication'' between the features across the track, we form a new feature matrix $\mathbf{Y}' \in \mathbb{R}^{T \times 2M}$, where each row contains the original feature vector in that row and an average feature vector for the whole track. Therefore $\mathbf{Y}' = \left[ \mathbf{y}'_1, \mathbf{y}'_2, \dots, \mathbf{y}'_T \right]^\top$, where
\begin{equation}
\mathbf{y}_\tau' = \left[ \begin{array}{c}
	\mathbf{y}_\tau\\
	\frac{1}{T}\sum_{i=1}^T \mathbf{y}_i\\
\end{array}\right].
\end{equation}
From these feature vectors concatenated with the average feature vector, we generate activations by another fully connected layer
$\mathbf{a}_\tau = \mathbf{W}_2\mathbf{y}'_\tau$, forming matrix $\mathbf{A} \in \mathbb{R}^{T \times M}$. These activations are then normalized by softmax; however, the normalization is not done by rows (as usually), but by columns to normalize the activation for each component of the feature vector. Therefore, the normalization yields matrix $\mathbf{E} \in \mathbb{R}^{T \times M}$, where
\begin{equation}
e_{\tau j} = \cfrac{\exp(a_{\tau j})}{\sum_{i=1}^T \exp(a_{ij})}.
\end{equation}

The weight matrix $\mathbf{E}$ is then merged with the compressed feature matrix $\mathbf{Y}$ by Hadamard (element-wise) product into matrix $\mathbf{Z} = \mathbf{Y} \circ \mathbf{E}$. The final feature vector $\mathbf{f}$ is then obtained as a sum of feature vectors in rows of matrix $\mathbf{Z}$, normalized to a unit vector.
\begin{equation}
\mathbf{f}  = \frac{\sum_{\tau=1}^T \mathbf{z}_\tau}{\left\| \sum_{\tau=1}^T \mathbf{z}_\tau \right\|_2}
\end{equation}
Therefore, if matrix $\mathbf{A}$ contained a single constant value, the aggregation would be reduced to one fully connected layer followed by average pooling. Instead, the weights $\mathbf{W}_1, \mathbf{W}_2, \mathbf{b}_1$ are trained by back-propagation and the network is thus able to produce better features.

\begin{figure}[t]
	\includegraphics[width=\linewidth]{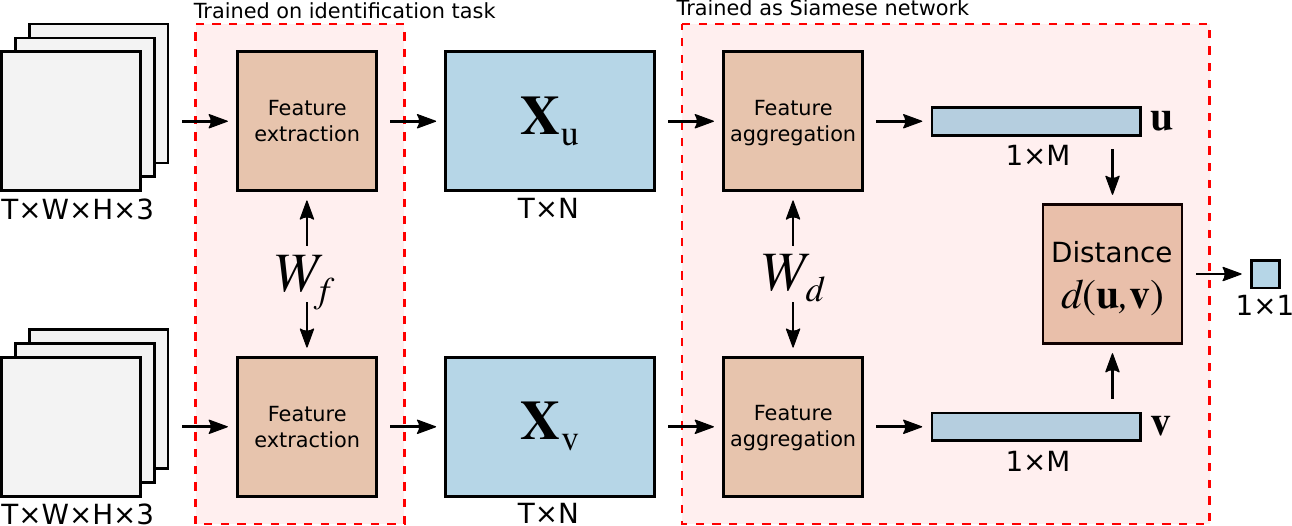}%
	\caption{Schematic design representing the full training and inference pipeline. In our approach, we train the image feature extractor NN on the identification task on the given dataset; however, the proposed method for feature aggregation can work with an arbitrary image feature extractor. \new{$W_f$ and $W_d$ refer to the shared weights of feature extractor part and feature aggregation part, respectively.}
	}
	\label{fig:SiameseNetwork}
\end{figure}

\subsection{Metrics for Distance Computation} \label{sec:MethodologyMetrics}

The re-identification task is defined by a query sample (track) and a gallery of samples (tracks), where one sample from the gallery is supposed to have the same identity as the query sample.  It is common to use Euclidean (or cosine for unit feature vectors) distance
$d_\mathrm{E}(\mathbf{u}, \mathbf{v}) = \sqrt{\sum_i (u_i - v_i)^2}$
to rank the gallery samples by their distance from the query feature vector.

Previous works have shown that other distance metrics can outperform the Euclidean one, and the Mahalanobis distance seems to be powerful \citep{Kostinger2012,Liao2015}.
Mahalanobis distance between vectors $\mathbf{u}$ and $\mathbf{v}$ is computed as $\sqrt{(\mathbf{u}-\mathbf{v})^\top \mathbf{M} (\mathbf{u}-\mathbf{v})}$, requiring that matrix $\mathbf{M}$ is symmetric and positive semi-definite \citep{Shi2016}. They claim that such a constraint is hard to enforce and propose to decompose the matrix $\mathbf{M} = \mathbf{W}\mathbf{W}^\top$ and learn $\mathbf{W}$ instead. Then, the Mahalanobis distance is computed by the following equation:
\begin{equation}
d_\mathrm{M}(\mathbf{u}, \mathbf{v}) = \sqrt{(\mathbf{u}-\mathbf{v})^\top \mathbf{W}\mathbf{W}^\top (\mathbf{u}-\mathbf{v})}.
\end{equation}

Using Mahalanobis distance as proposed by \cite{Shi2016} improves the re-identification accuracy, paying a high price in terms of its time complexity. Both time and memory asymptotic complexities are $\mathcal{O}(D^2)$ where $D$ is the dimensionality of the feature vectors.
This can cause significant problems in re-identification as the computational cost for quadratic time complexity is significantly larger even for $D=128$. Therefore, we propose to learn suitable weights for Weighted Euclidean distance (equivalent to Mahalanobis distance when matrix $\mathbf{M}$ is diagonal), instead. We express the Weighted Euclidean distance by
\begin{equation}
d_\mathrm{WE}(\mathbf{u}, \mathbf{v}) = \sqrt{\sum_{i=1}^D w_i (u_i - v_i)^2},
\end{equation}
where $\mathbf{w} = [w_1, w_2, \dots, w_D]$ are learned weights. It should be noted that if all the weights $w_i$ are equal to $1$, the metric is reduced to standard Euclidean distance. Before learning, we initialize the weights by randomly sampling from normal distribution with $\mu = 1$ and $\sigma = 0.1$.

As the Weighted Euclidean distance can be interpreted as Mahalanobis distance with diagonal matrix $\mathbf{M}$, the same conditions must be kept. The symmetricity is satisfied trivially as it is a diagonal matrix. However, to ensure the positive semi-definite property, we ensure that all the weights $w_i$ are non-negative by clipping values bellow zero after each update of the weights during learning.

The Weighted Euclidean distance has benefits when compared to both standard Euclidean and Mahalanobis distances. Compared to the Euclidean distance, it has a higher expressive power thanks to learned weights $\mathbf{w}$. On the other hand, compared to full Mahalanobis distance, it is much faster as both time and memory complexity of the Weighted Euclidean distance is $\mathcal{O}(D)$. At the same time, as the results in Section~\ref{sec:ExperimentsVehicle} show, our proposed Weighted Euclidean distance also outperforms both Euclidean and full Mahalanobis distance in terms of re-identification accuracy.

\subsection{Full Training and Inference Network}

Both the feature aggregation network (Section \ref{sec:MethodologyFeatureAggregation}) and the Weighted Euclidean metric (Section \ref{sec:MethodologyMetrics}) are trained by a Siamese network \citep{Hadsell2006}, see Figure~\ref{fig:SiameseNetwork}. For speeding up the training, we pre-train the feature extractor (Inception-ResNet-v1 \citep{Szegedy2017inception} for vehicle re-id and ResNet50 \citep{He2016} for person in our case) for the identification task using the dataset training data and then we cache all features for the tracks and train the feature aggregation and distance metric with the cached features. Training the network end-to-end did not improve the results further.
We use a standard contrastive loss \citep{Hadsell2006}
\begin{equation}
L(\mathbf{u}, \mathbf{v}, y) = y \cdot d(\mathbf{u}, \mathbf{v})^2 + (1-y) \cdot \left[m - d(\mathbf{u}, \mathbf{v})\right]^2_+,  \label{eq:ContrastiveLoss}
\end{equation}
where $\mathbf{u}$ and $\mathbf{v}$ are feature vectors, $m$ is the margin between negative samples, $\left[\dots\right]_+$ denotes maximum value with zero, and $y=1$ if $l_u=l_v$ or $0$ otherwise ($l_u$ and $l_v$ are sample identities). Distance $d$ is one of $d_\mathrm{E}$, $d_\mathrm{M}$, or $d_\mathrm{WE}$ from the previous section.

\begin{figure}[t]
	\centering
	\includegraphics[width=\linewidth]{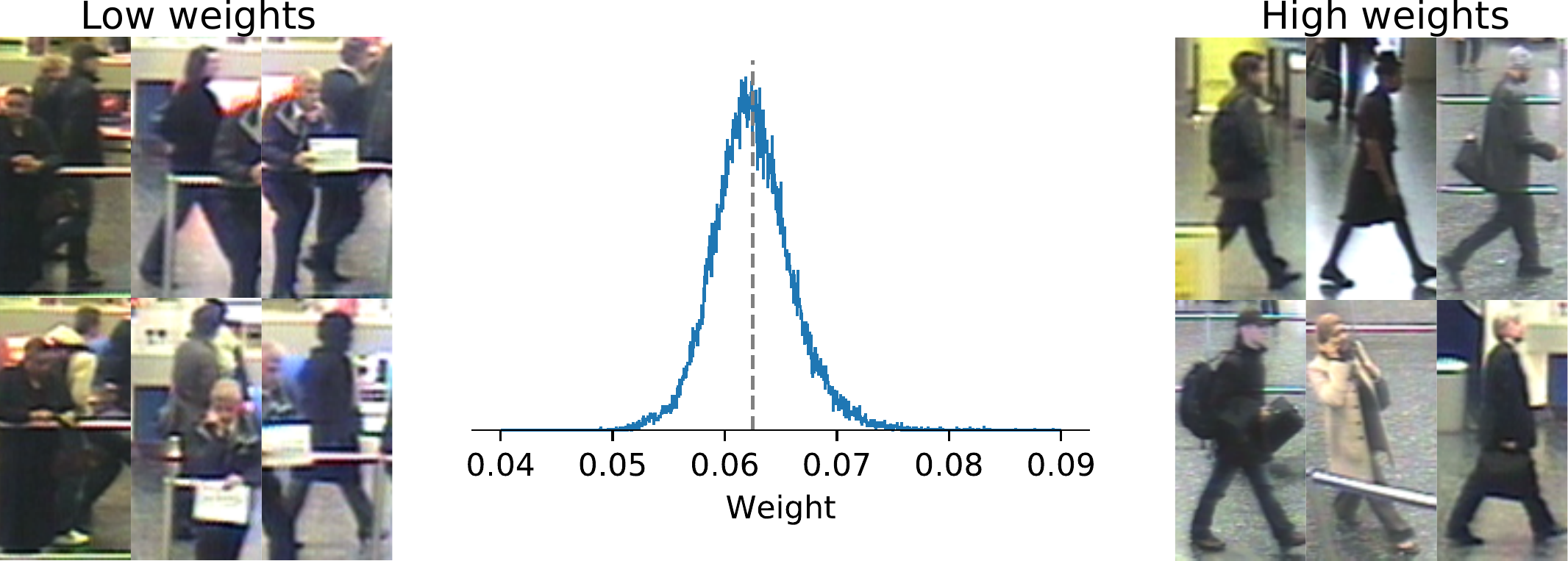}%
	\caption{\textbf{Middle:} Distribution of mean weights for test images in iLIDS-VID dataset \citep{Wang2014}. The dashed grey line denotes image weight for average pooling with $T=16$. \textbf{Sides:} Images with the lowest and highest weights which show that low weight is usually assigned to images with occluding pedestrians.}
	\label{fig:FrameWeights}
\end{figure}

\subsection{Design Choices}
We analyzed several design choices we made. During preliminary experiments we used ReLU nonlinearity in Equation~\eqref{eq:projection} and found out that the results are significantly better with $\tanh$ nonlinearity.

\begin{figure}[t]
	\centering
	\includegraphics[width=\linewidth]{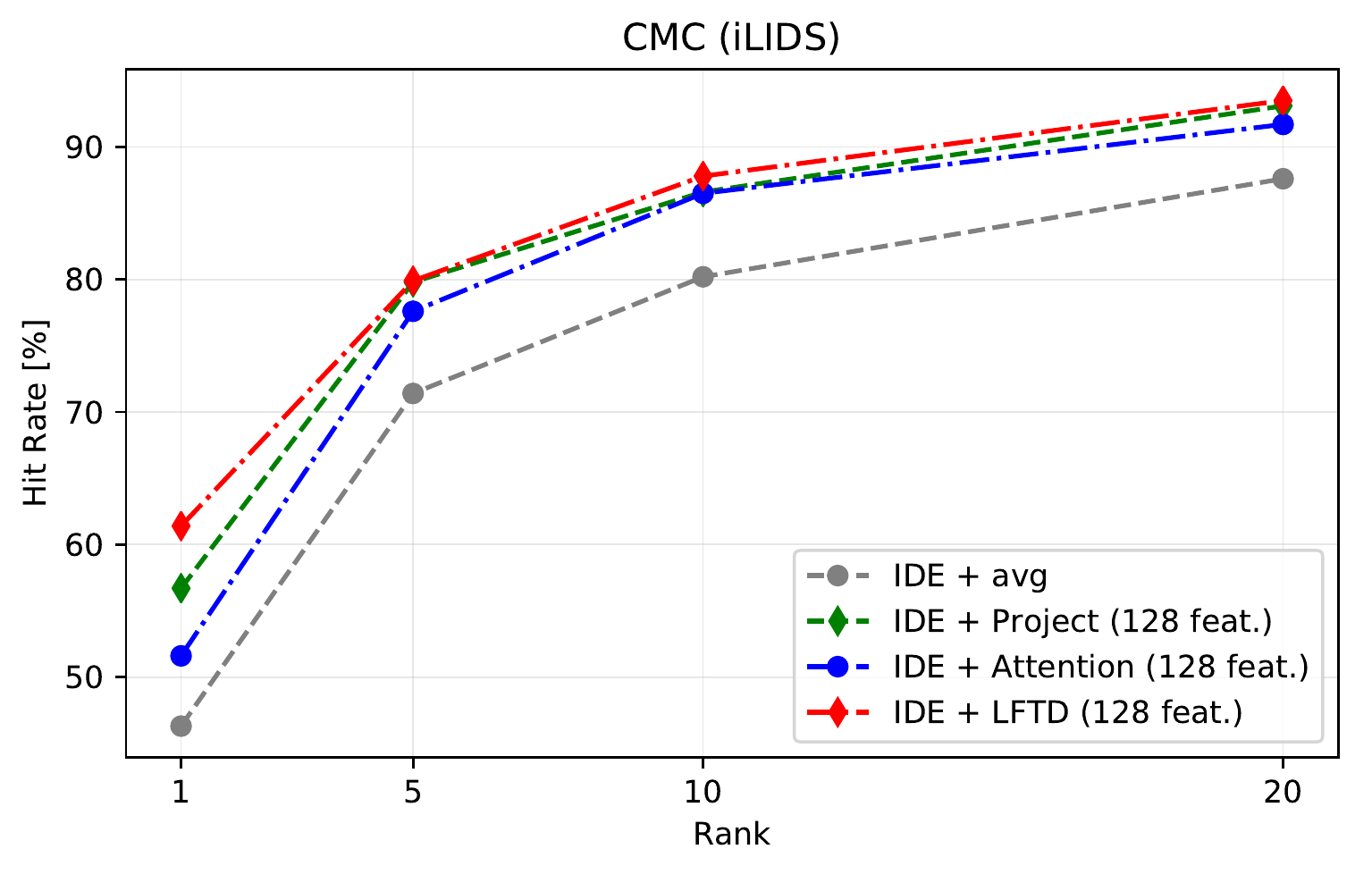}%
	\caption{\new{CMC curve on the iLIDS dataset for individual parts of the proposed network.}}
	\label{fig:DesignChoicesILIDS}
\end{figure}

Furthermore, on iLIDS-VID dataset \citep{Wang2014}, we tested how important different parts of the network are. In these experiments, 128 dimensional features were used (except the average pooling, where the features had 2048 dimensions). When only average pooling was used, we got Hit@1 $46.3\,\%$ and with the full network Hit@1 is $61.4\,\%$. However, if we use only the weighting mechanism (omit feature projection by \eqref{eq:projection}), the Hit@1 is $51.6\,\%$. And finally, if we use average pooling (omit the weighting mechanism) with the feature projection \eqref{eq:projection}, we receive Hit@1 $56.7\,\%$. This shows that both parts of the network contribute to the accuracy and the contributions can be merged to obtain better results. \new{A graphical comparison of design choices evaluation can be found in Figure \ref{fig:DesignChoicesILIDS}.} Full results of design choices evaluation for different Hit@Rank can be found in Table \ref{tab:DesignChoicesILIDS}.

Finally, we analyzed the mean weights for different images and the distribution of mean weights together with images with lowest and highest weights can be found in Figure \ref{fig:FrameWeights}. The results show that the weights are centered around $1/T$ (i.e. average pooling weight)  which was expected. Also, low weights are usually assigned to images with occluding objects or pedestrians.

Furthermore, we analyzed the homogeneity of the weights for individual observations (i.e. how much the weights differ within one observation). The mean relative standard deviation is 0.34; the weights therefore differ significantly.


\begin{table}
	\centering
	\scriptsize
	\caption{Evaluation of individual parts of the proposed network on the iLIDS dataset.}
	\label{tab:DesignChoicesILIDS}
	\begin{tabular}{lrrrr}
		\toprule
\textbf{Method} & \multicolumn{4}{c}{\textbf{Hit@R (iLIDS)}}\\
									   &    1  &    5  &   10 &   20 \\
\midrule
		IDE + avg                      & 46.3  & 71.4  & 80.2 & 87.6 \\
		IDE + Project (128 feat.)      & 56.7  & 79.8  & 86.6 & 93.1 \\
		IDE + Attention (128 feat.)    & 51.6  & 77.6  & 86.5 & 91.7 \\
		IDE + LFTD (128 feat.)         & 61.4  & 79.9  & 87.8 & 93.5 \\
		\bottomrule
	\end{tabular}
\end{table}


%% file: 2018-CVIU-FeatureImportance-4-Dataset.tex
\section{Novel Vehicle Re-Identification Dataset \dataset}
\label{sec:Dataset}

\begin{table}[t]
	\scriptsize
	\centering
	\caption{The comparison of various vehicle re-id datasets.
		\newline * -- Tracks are not guaranteed for each unique vehicle.
		\newline $\dagger$ -- Unique vehicles from each dataset part can overlap. The total number of unique vehicles is probably lower.
	} 
	\begin{tabular}[t]{l r r r r}
		\toprule
		& CarsReId74k & VehicleID & VeRi-776 & PKU-VDs \\
		\midrule
		\# unique vehicles & 17,681 & 26,267 & 776 & \textbf{$^\dagger$221,519} \\
		\# tracks & \textbf{73,976} & --- & 6,822 &  N/A\\
		\# images & \textbf{3,242,713} & 221,763 & 51,035 & 1,354,876 \\
		viewpoints & \textbf{various} & front/rear & \textbf{various}  & front \\
		multiple images in track & \textbf{yes} & no & \textbf{*yes}  &  \textbf{yes}\\
		\bottomrule
	\end{tabular}
	
	\label{tab:DatasetsComp}
\end{table}


We focus on vehicle re-identification and we want to differentiate even vehicles with the same fine-grained type but different identities (different license plates). Therefore, we cannot use fine-grained vehicle recognition datasets~\citep{Sochor2016BoxCars,Sochor2017BoxCars,Yang2015,Krause2013} for the task.
As other existing vehicle re-identification datasets VeRi-776 \citep{Liu2016deep}, VehicleID \citep{Liu2016CVPR} and PKU-VDs \citep{Yan2017} are either small (VeRi-776) or limited to frontal/rear viewpoints (VehicleID, PKU-VDs). We collected a novel dataset \textbf{\dataset} which does not have these limitations. The data were collected by \camerasAll\ cameras from various angles and the dataset contains almost 74\,k of vehicle tracks with precise identity annotation (acquired from license plates). \new{More detailed comparison of different available vehicle re-identification datasets can be found in Table \ref{tab:DatasetsComp}.}

\begin{figure*}[t]
	\centering
	\tmpframe{\includegraphics[width=0.5\linewidth]{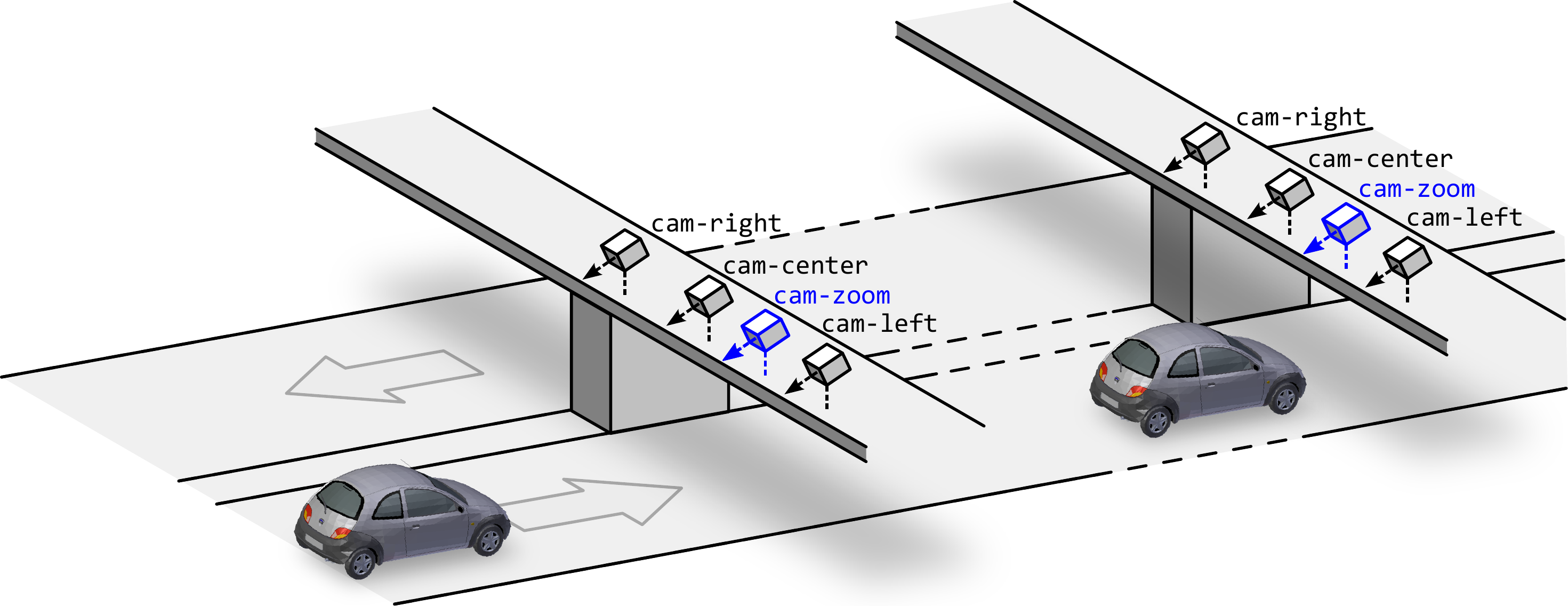}}
	\tmpframe{\includegraphics[width=0.4\linewidth]{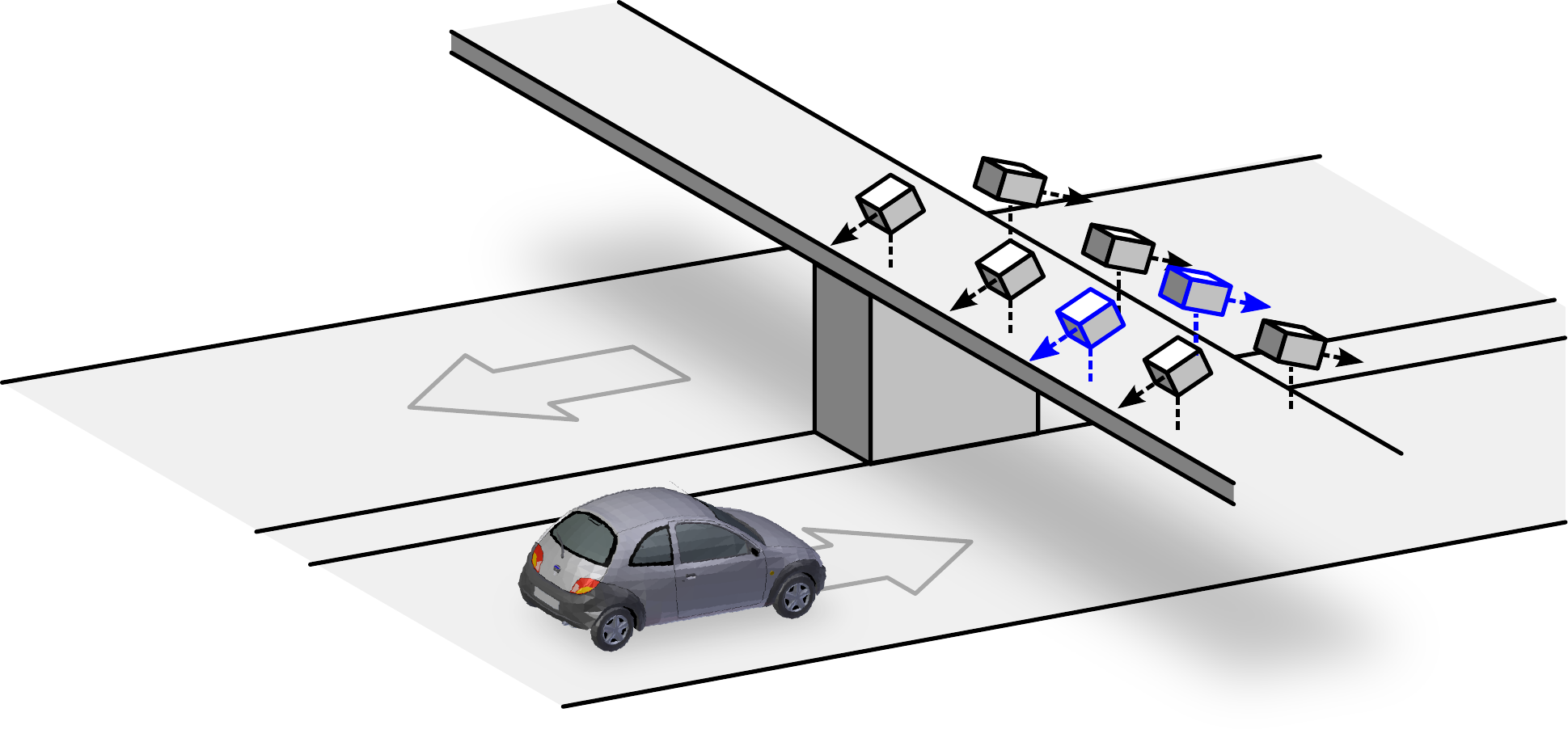}}
	\caption{Recording setup for acquisition of novel \dataset\ dataset. We simultaneously recorded data on two bridges by multiple cameras. One camera on each bridge was zoomed in so that it is possible to automatically recognize license plates and use them for the construction of the ground truth labeling \new{(left image)}.
	For part of dataset vehicles were captured from single bridge on both sides, which yields to capture observed vehicles from frontal and rear viewpoints \new{(right image)}.
	}
	\label{fig:RecordingSetup}
\end{figure*}

\subsection{Dataset Acquisition}
\label{sec:DatasetAcquisition}

\begin{figure}[t]
	\fbox{\includegraphics[width=0.25\linewidth]{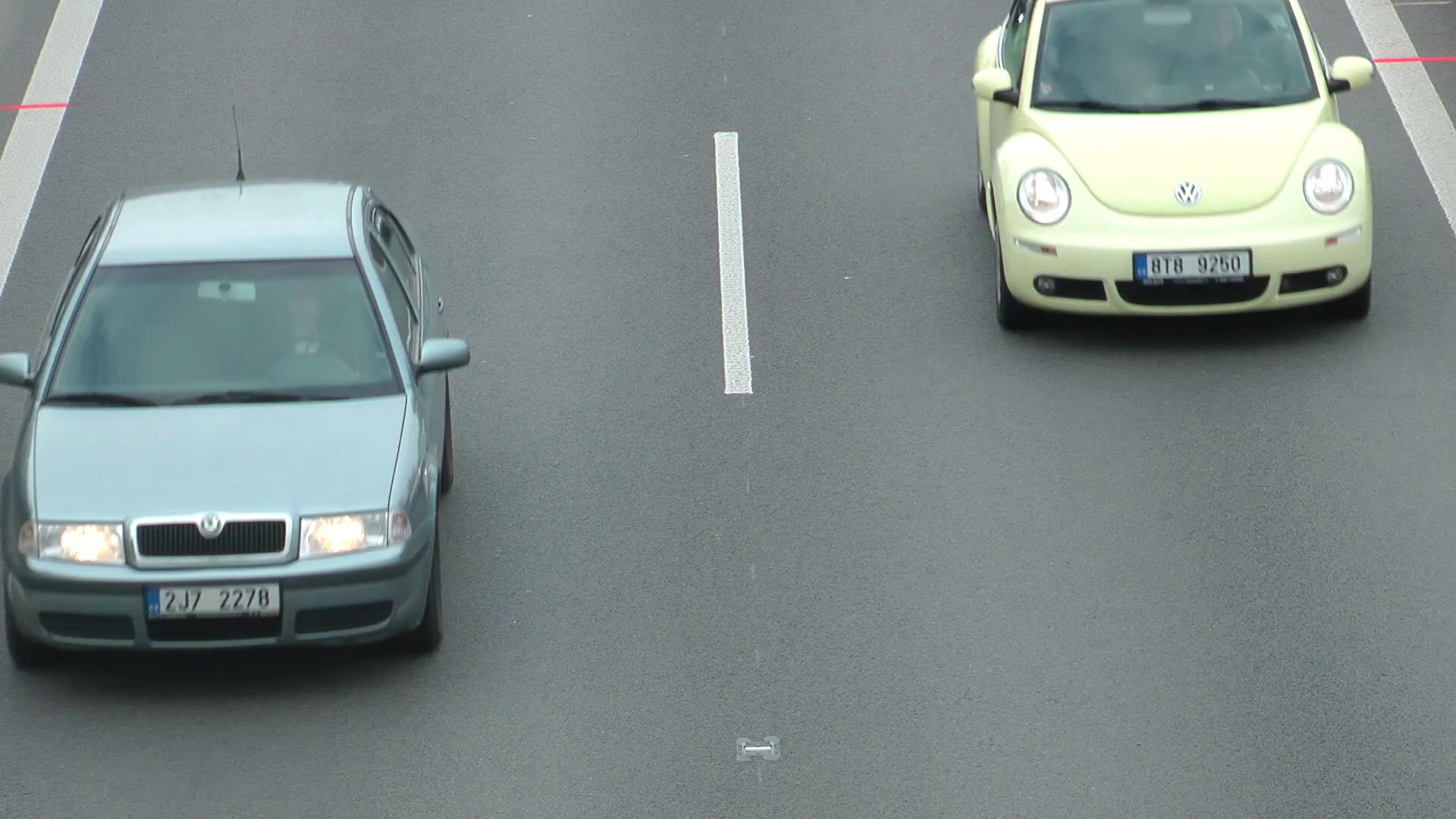}%
		\includegraphics[width=0.25\linewidth]{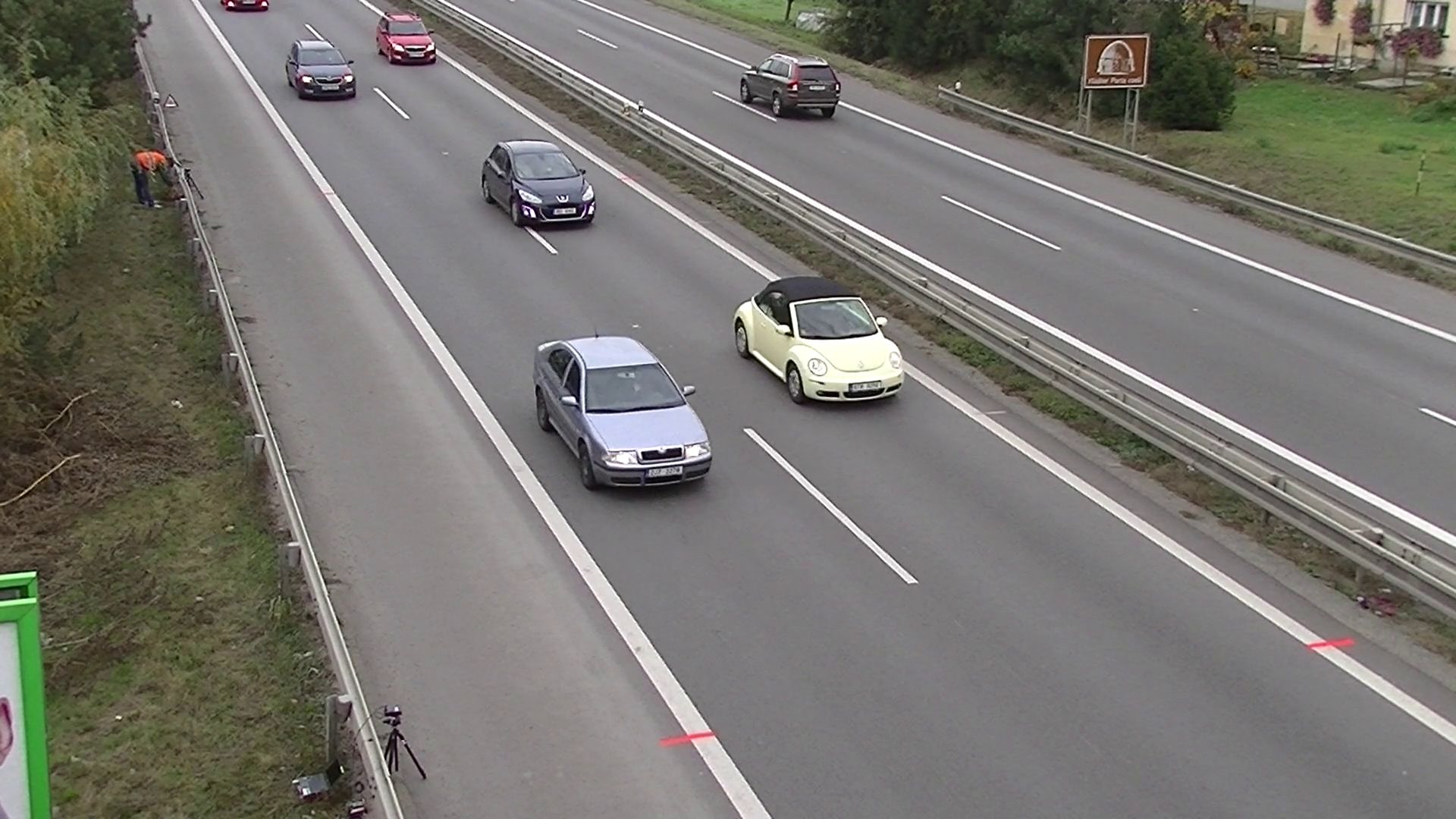}%
		\includegraphics[width=0.25\linewidth]{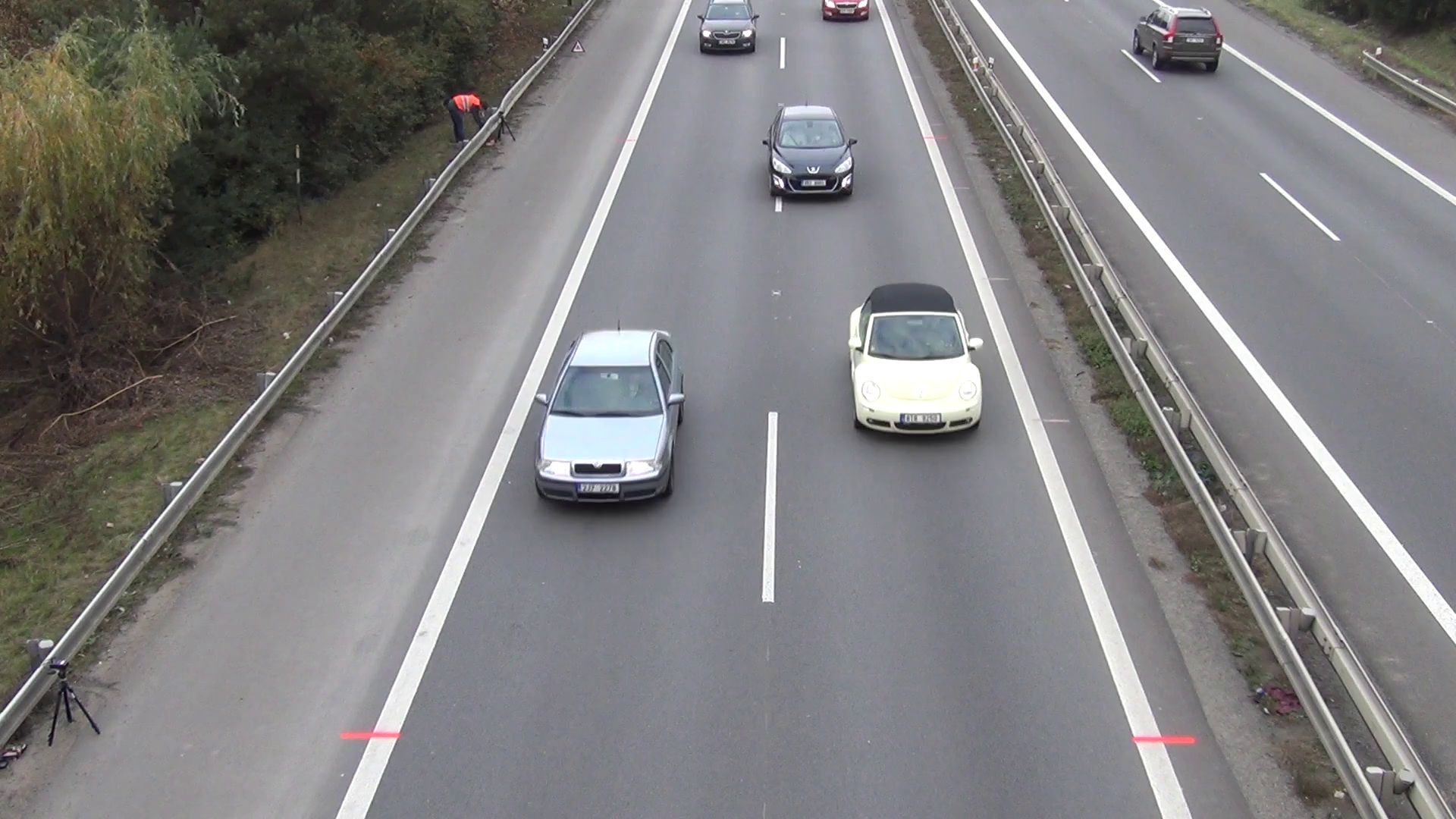}%
		\includegraphics[width=0.25\linewidth]{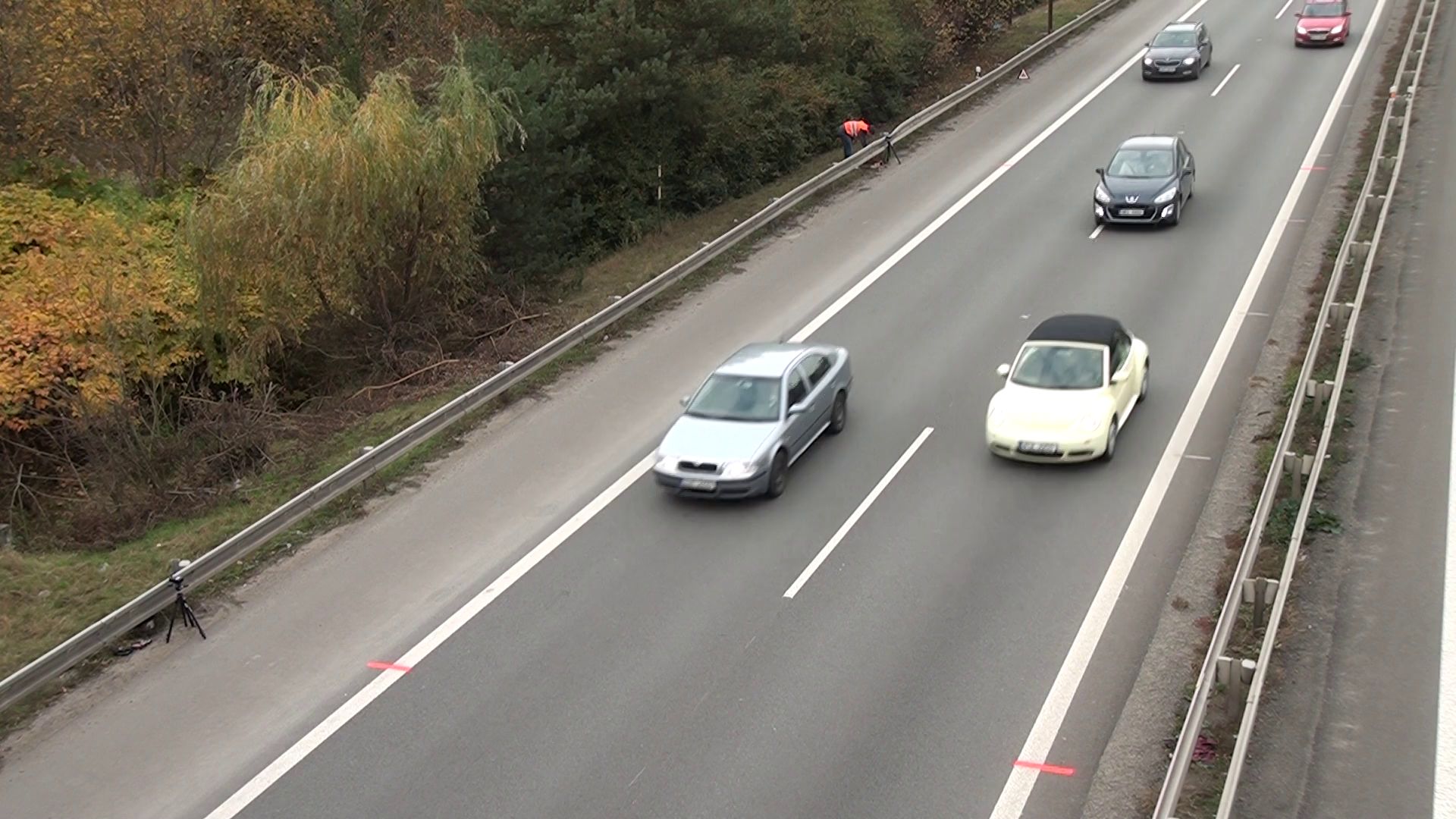}}\\[-2pt]
	\fbox{\includegraphics[width=0.25\linewidth]{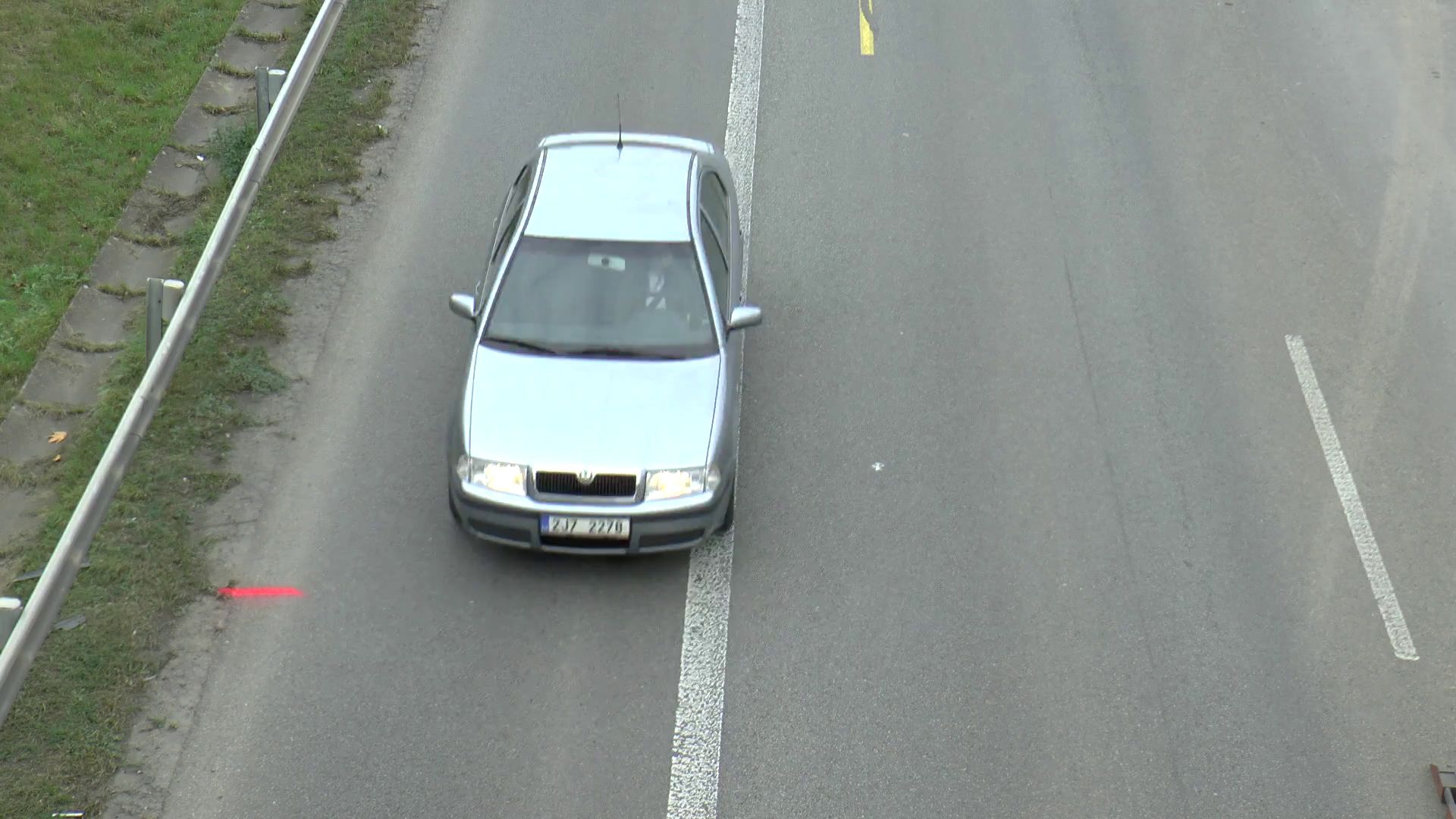}%
		\includegraphics[width=0.25\linewidth]{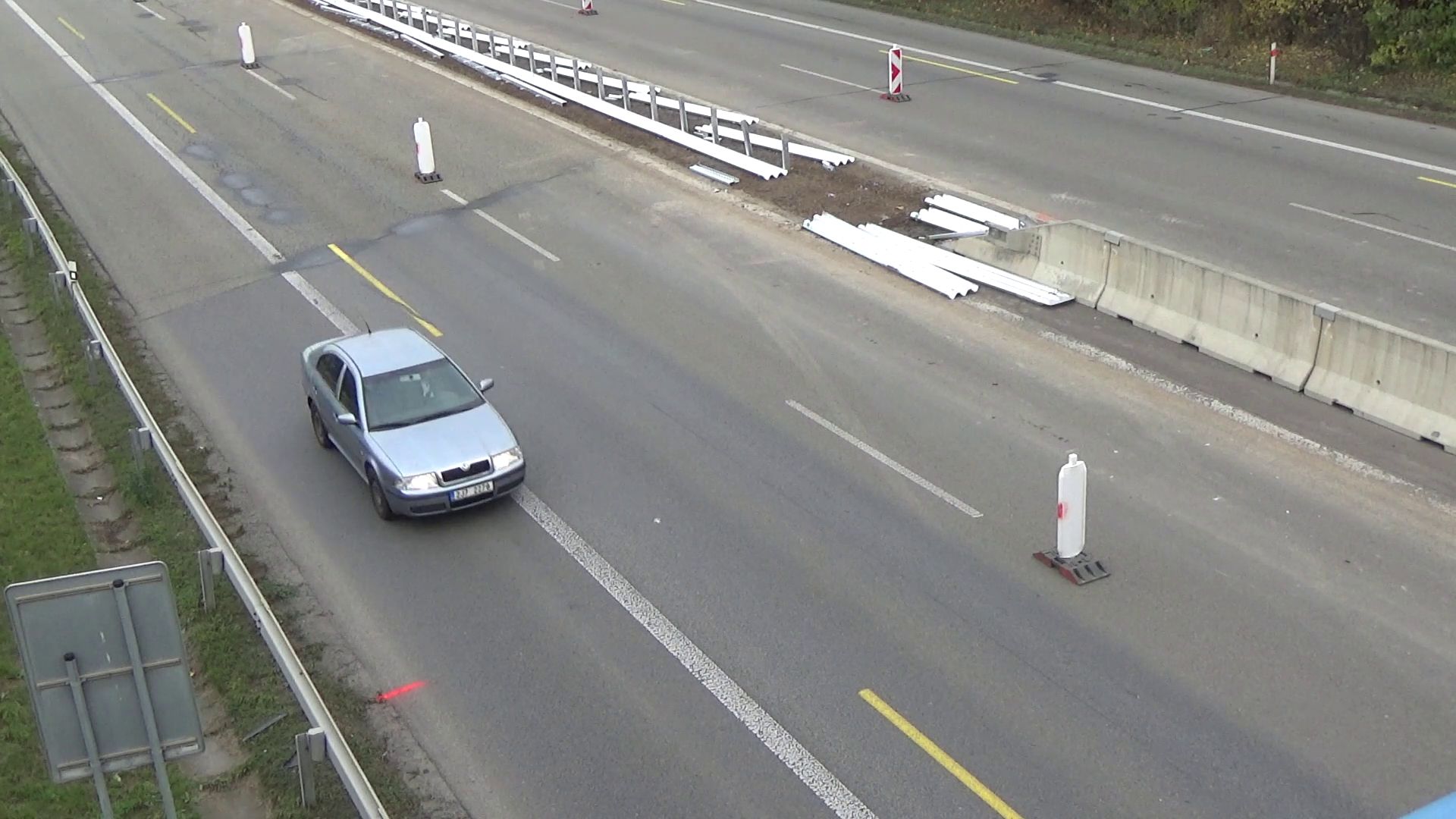}%
		\includegraphics[width=0.25\linewidth]{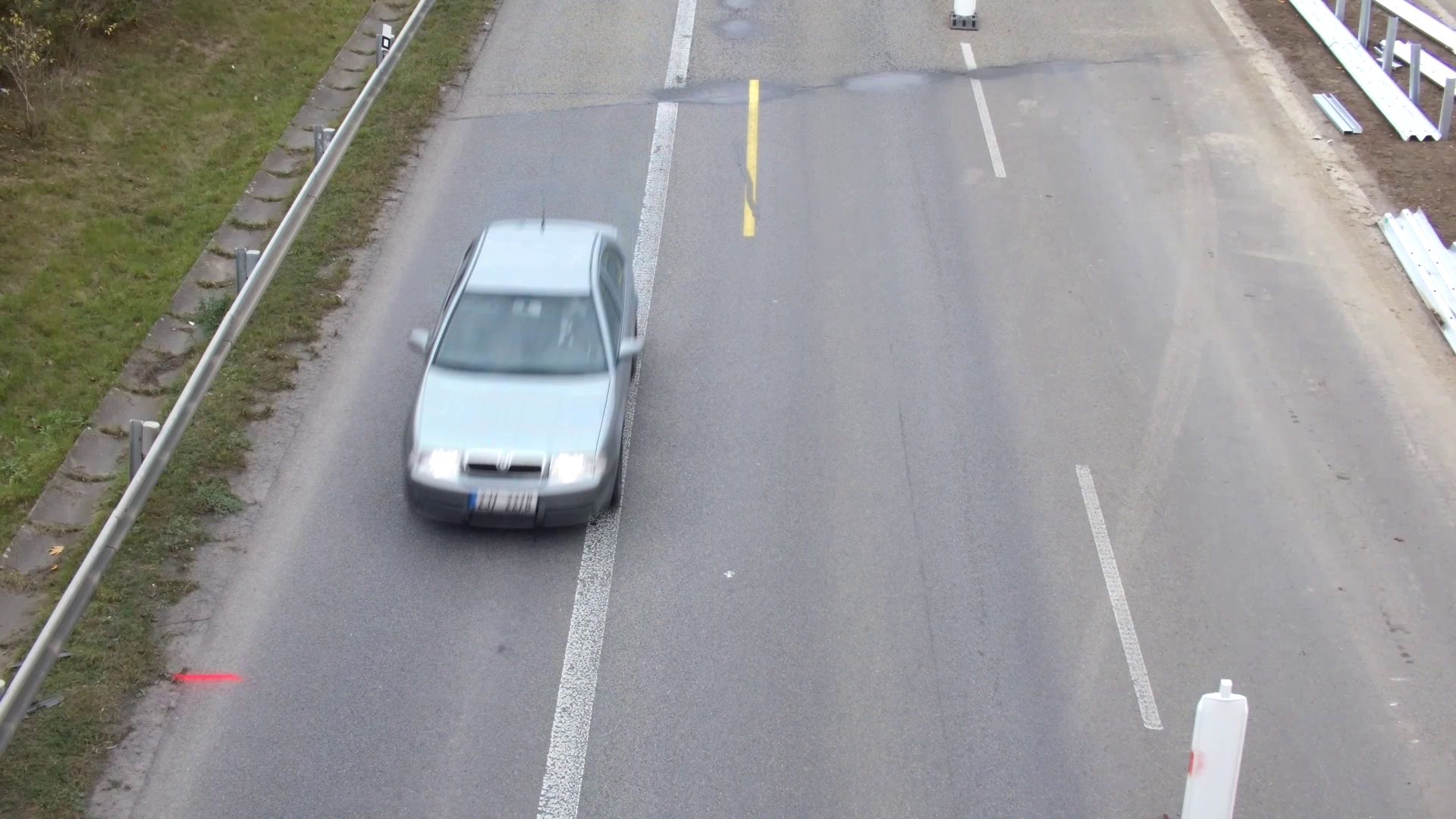}%
		\includegraphics[width=0.25\linewidth]{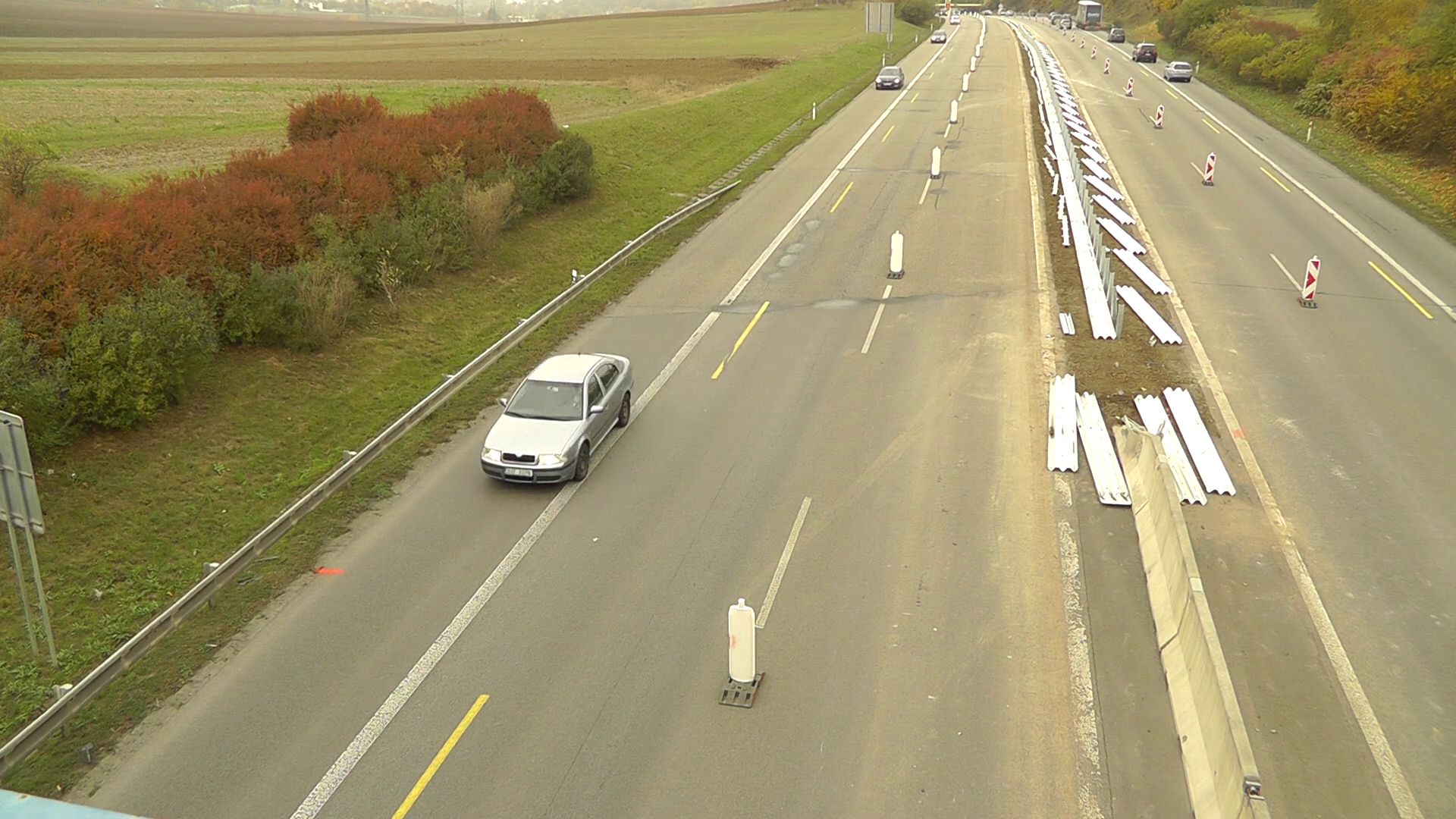}}
	\caption{Frames from all cameras in one session. The license plates acquired from the \emph{zoom} camera (left) were used for ground truth re-identification (silver car). Each row shows frames from one location within the session.
	}
    \label{fig:Recording}
\end{figure}

The dataset was collected in multiple sessions. In each session, we placed four cameras on a bridge overlooking a freeway and four cameras on another bridge in vehicles' traveling direction. Figure~\ref{fig:RecordingSetup} illustrates the recording setup and Figure~\ref{fig:Recording} shows example frames from one such session. The videos were recorded for $\sim1$ hour and synchronized. One of the cameras was zoomed in enough to be able to read the license plates of all the passing vehicles (Figure~\ref{fig:Recording} left). The other three cameras were placed so that they observed the road from left, center, and right position.

\begin{figure}[t]
	\definecolor{querycolor}{HTML}{1F77B4}
	\definecolor{poscolor}{HTML}{2CA02C}
	\definecolor{negcolor}{HTML}{D62728}
	\includegraphics[width=\linewidth]{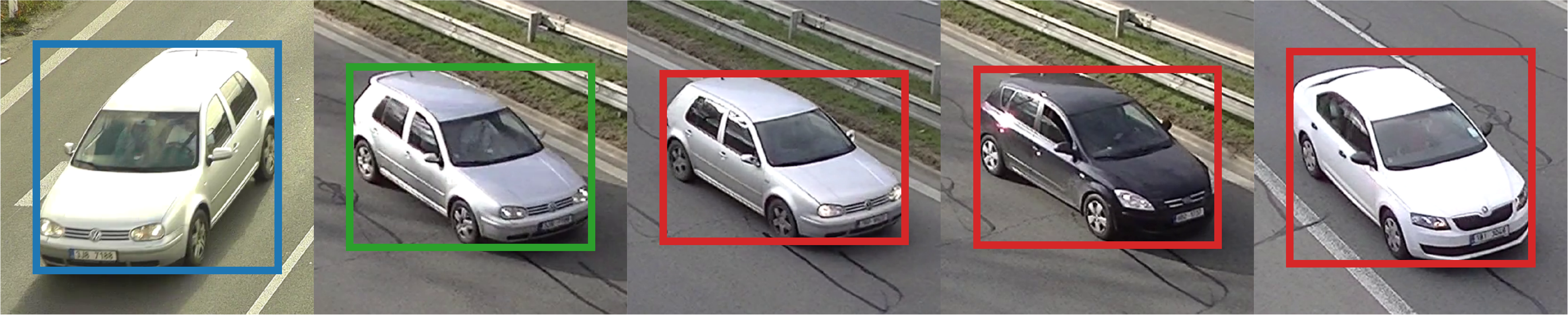}\\
	\includegraphics[width=\linewidth]{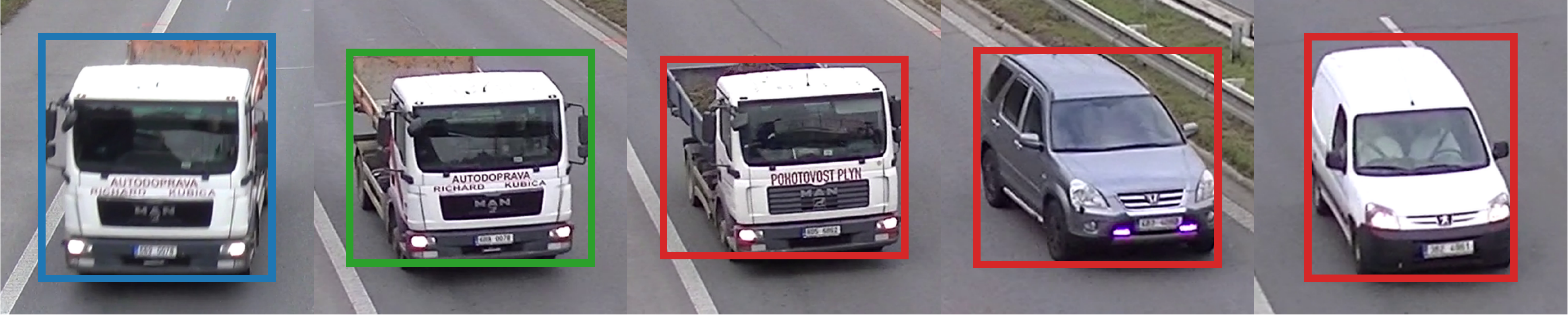}\\
	\includegraphics[width=\linewidth]{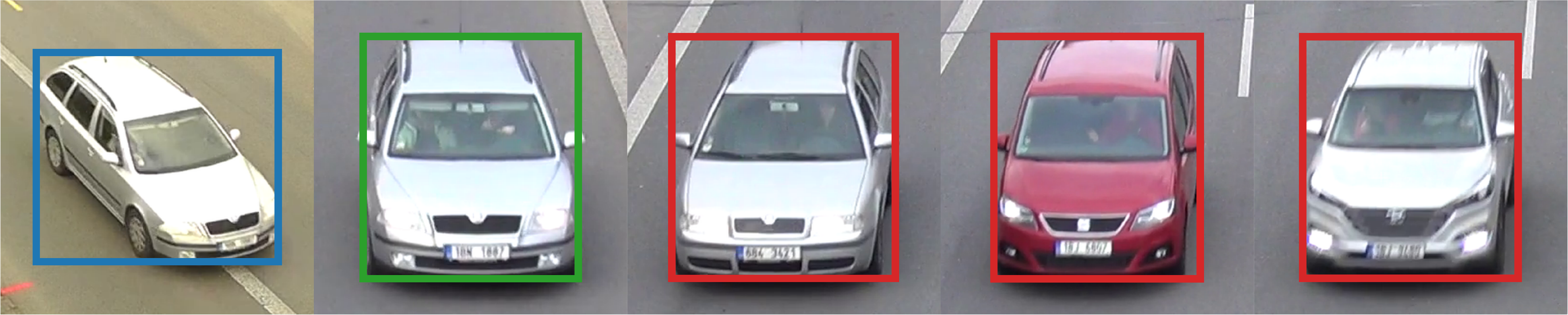}
	\caption{Examples of \textcolor{querycolor}{queries}, \textcolor{poscolor}{positive}, and \textcolor{negcolor}{negative} samples. The negatives are sorted by difficulty from left to right (hard to easy) based on distances obtained from our re-identification feature vectors. It should be noted that the hardest negative sample has usually subtle differences (e.g. missing a small spoiler in the first row).}
	\label{fig:DatasetPositiveNegative}
\end{figure}

We used the zoomed-in videos to identify the passing vehicles. We detected the license plates by an ACF detector~\citep{Dollar2014}, tracked them, recognized by a recent method~\citep{Spanhel2017holistic} and manually verified in order to eliminate any recognition errors. 
We also assigned a lane to each  license plate track for easier matching.
On all the other videos (left, center, right), we detected
and tracked the vehicles. We also constructed 3D~bounding boxes~\citep{Dubska2014} around the vehicles as \cite{Sochor2016BoxCars} showed that the 3D~bounding boxes were beneficial for fine-grained recognition. We also assigned the lane for each of these vehicles~\citep{Dubska2014}.
Finally, we matched the vehicles from the zoomed-in cameras (with known identities) to vehicles from the other cameras. We omitted all the vehicles which were not matched. 
It should be noted that the vehicles in the dataset from non-zoomed-in cameras have almost unreadable license plates; therefore, the dataset is suitable for \textbf{appearance-based} vehicle re-identification, preserving the anonymity of the vehicles.

\subsection{Dataset Statistics}
The dataset was recorded in \new{11} sessions at different locations. We divided the dataset into \new{the training, the testing and the validation part by sessions (five sessions for training, five sessions for testing and one validation)}. The total dataset statistics can be found in Table~\ref{tab:DatasetStats}. The table shows that our dataset is significantly larger than VeRi-776 \citep{Liu2016deep} dataset with only 776 unique vehicles. And compared to VehicleID, VD1 and VD2 datasets \citep{Liu2016CVPR,Yan2017}, our dataset is not limited to frontal/rear viewpoints.
Compared to VehicleID dataset, \dataset\ dataset has fewer unique vehicles (17,681~vs.~26,267), however far more image (\imagesAll\ vs. 221,763) as vehicles are seen from more viewpoints. 

\subsection{Proposed Evaluation Protocol}
For each part (training, testing and validation), we collected all the pairs of tracks with the same vehicle identity (marked as \textit{query}, \textit{positive}). The query and positive tracks are always from different videos; however, they can come from the same session and location (e.g. left -- right), from the same session and different location, or (in rare cases) also from different sessions within the training (or testing) set. This yields a significant number of positive pairs (\pairsAll\ in total).
As the negative pairs, we use all other vehicle tracks in the same video as the positive track with the exception of vehicle tracks with the same identity as the positive track (a vehicle could be observed multiple times in one video). 
This yields a mean number of \negativesAll\ negative vehicle tracks per positive pair.
See Figure~\ref{fig:DatasetPositiveNegative} for examples of positive and negative pairs.

Following other papers \citep{Liu2016CVPR,Yan2017,Liu2016deep,Hirzer2011,Wang2014} on re-identification, we use \textbf{mAP} and \textbf{hit at rank} as the metrics for evaluation on the dataset. We encourage others to report hit rates at ranks 1, 5, 10, and 20 together with Cumulative Matching Curve for ranks 1 to 20.

%% file: 2018-CVIU-FeatureImportance-5-Results.tex
\section{Experimental Results}
\label{sec:Experiments}
We evaluate our method on the vehicle and person re-identification tasks on multiple public datasets to show that the aggregation performs well on various classes of data.
Datasets for evaluation were chosen considering the availability of tracks (multiple observations) of each object's identity in the dataset because this work proposes a method for aggregation of features in the time domain \new{and variable camera viewpoints.}
\subsection{Vehicle Re-Identification}
\label{sec:ExperimentsVehicle}

\new{Currently available datasets does not fit conditions described before at least in one condition (see Sec.~\ref{sec:Dataset}), thus vehicle re-identification task was evaluated on our novel \dataset\ only.}

For feature extraction from images we use \textbf{Inception-ResNet-v2} \citep{Szegedy2017inception} with images resized to $331\times331$ yielding feature vectors with length \avgFeatCars\ for each input image. \cite{Sochor2016BoxCars,Sochor2017BoxCars} showed that unpacking the input vehicle by 3D bounding box and alternating the input image colors is beneficial for fine-grained recognition of vehicles; we use these modifications for re-identification of vehicles as well.

The feature extractor was fine-tuned on the identification task using the training part of the \dataset\ dataset. The fine-tuning was done with Adam optimizer, learning rate 0.0001, batch size 4 for 300 epochs with standard augmentation techniques (random flip and shift of the bounding box).

\begin{table}[t]
	\centering
	\scriptsize
	\caption{\dataset\ dataset statistics. $^*$The total number of unique vehicles is lower than the sum of unique vehicles from training, test and validation set because a small number of vehicles appear in all sets (same car present at two or more recording sessions by accident). $^\dag$Number of negative pairs = mean number of negative pairs per positive pair.}
	\label{tab:DatasetStats}
	\begin{tabular}{l r r r r}
		\toprule
								& training 		& test 			& validation	& total\\
		\midrule
		\# cameras 				& 30 			& 30			& 6 			& \camerasAll \\
		\# unique vehicles$^*$ 	& 7,658 		& 9,678			& 1,100			& \vehiclesAll \\
		\# tracks 				& 32,163  		& 36,535 		& 5,278			& \tracksAll \\
		\# images 				& 1,469,494		& 1,467,680 	& 305,539 			& \imagesAll \\
		\midrule
		\# positive pairs 		& 125,086 		& 129,774 		& 22,376			& \pairsAll \\
		\# negative pairs$^\dag$ & 1,149 		& 1,459 		& 881			& \negativesAll \\
		\bottomrule
	\end{tabular}
\end{table}


\begin{table}[t]
	\centering
	\scriptsize
	\caption{Results for different methods for vehicle re-identification on \dataset\ dataset. The methods use 128 dimensional feature vectors with the exception of avg which uses \avgFeatCars\ dimensional feature vectors. The methods use Euclidean distance with the exception of LFTD -- M (full Mahalanobis \citep{Shi2016}) and LFTD -- WE (Weighted Euclidean as proposed in Section~\ref{sec:MethodologyMetrics}).
		\emph{Input modifiers} -- UNP, UNP+IM \citep{Sochor2017BoxCars}. \emph{Aggregation methods} -- RNN \citep{McLaughlin2016}, NAN \citep{Yang2017}.}
	\label{tab:VehicleReIdResults}
	\begin{tabular}{llrrrrr}
		\toprule
		& && \multicolumn{4}{c}{Hit@Rank} \\
		Input Modif.   & Aggregation             & mAP   & 1   & 5   & 10   & 20   \\
		\midrule
		None   & avg               & 0.608 & 55.3    & 66.4    & 71.3     & 76.5     \\
		UNP    & avg               & 0.652 & 58.4    & 72.8    & 78.0     & 83.1     \\
		UNP+IM & avg               & 0.672 & 61.2    & 73.8    & 78.7     & 83.5     \\
		\midrule
		UNP+IM & RNN               & 0.678 & 59.0    & 78.2    & 84.5     & 89.7     \\
		UNP+IM & NAN               & 0.700 & 63.3    & 77.5    & 82.7     & 87.5     \\
		\midrule
		UNP+IM & LFTD              & 0.746 & 68.5    & 81.6    & 85.8     & 89.6     \\
		UNP+IM & LFTD -- M         & 0.757 & 69.5    & 83.2    & 87.3     & 90.7     \\
		UNP+IM & LFTD -- WE  & \textbf{0.779} & \textbf{71.3}    & \textbf{85.8}    & \textbf{89.9}     & \textbf{93.1}     \\
		\bottomrule
	\end{tabular}
\end{table}

When it comes to feature aggregation in temporal domain, we compare several methods with the following naming conventions:
\begin{itemize}[noitemsep,topsep=2pt]
	\item \textbf{avg} -- standard average pooling of feature vectors,
	\item \textbf{RNN} -- method proposed by \cite{McLaughlin2016} based on recurrent neural network,
	\item \textbf{NAN} -- \textbf{N}eural \textbf{A}ggregation \textbf{N}etwork proposed by \cite{Yang2017},
	\item \textbf{LFTD} -- our method (short for \textbf{L}earning \textbf{F}eatures in \textbf{T}emporal \textbf{D}omain).
\end{itemize}

To make the comparison fair, we always compare the methods with features of the same length (128 dimensional features by default). The only exception is average pooling where the final features are always \avgFeatCars\ dimensional. As NAN \citep{Yang2017} does not reduce the number of features, we added a trainable fully connected layer between the feature extractor and the aggregation network. As both RNN \citep{McLaughlin2016} and NAN~\citep{Yan2017} use Euclidean distance in the original design, we evaluate the networks with the Euclidean distance. Following other previous works \citep{McLaughlin2016,Zhang2017,Chen2017,Xu2017,Zhou2017}, we fix the number of time samples to $T=16$.

We also compare different metrics for comparison of the feature vectors. The standard Euclidean distance is used as the baseline. We also use the full Mahalanobis distance (as proposed by \cite{Shi2016}) -- shortened as \textbf{M}; and our Weighted Euclidean distance -- shortened as \textbf{WE}. The full Mahalanobis distance was trained with regularization term $0.5 \lambda \|\mathbf{W}\mathbf{W}^\top - \mathbf{I} \|_F^2$ as proposed by the authors \citep{Shi2016} with $\lambda = 0.01$.

To increase the training speed, all the aggregation networks were trained on cached features extracted by the Inception-ResNet-v2 feature extractor. The networks were trained in Siamese settings for 30 epochs with batch size 32 on train and validation set. We employed hard negative mi\-ning during the training and all positive pairs and one hardest negative pair per positive pair were presented to the network in one epoch during the training.
For the RNN \citep{McLaughlin2016},  we used original hyperparameters as proposed in the paper (SGD, lr: 0.001, margin: 2); changing them did not improve the accuracy further. We were forced to change the hyperparemters for NAN \citep{Yang2017} to different values than used in the paper as the network did not converge with the original ones. We used RMSprop optimizer with learning rate 1e-6, and margin 1; different hyperparameters did not improve the accuracy further. Our method LFTD was trained by Adam optimizer with learning rate 1e-5 (1e-4.4 in the case of Mahalanobis and Weighted Euclidean distance) and margin 2.


\begin{figure}[t]
	\includegraphics[width=\linewidth]{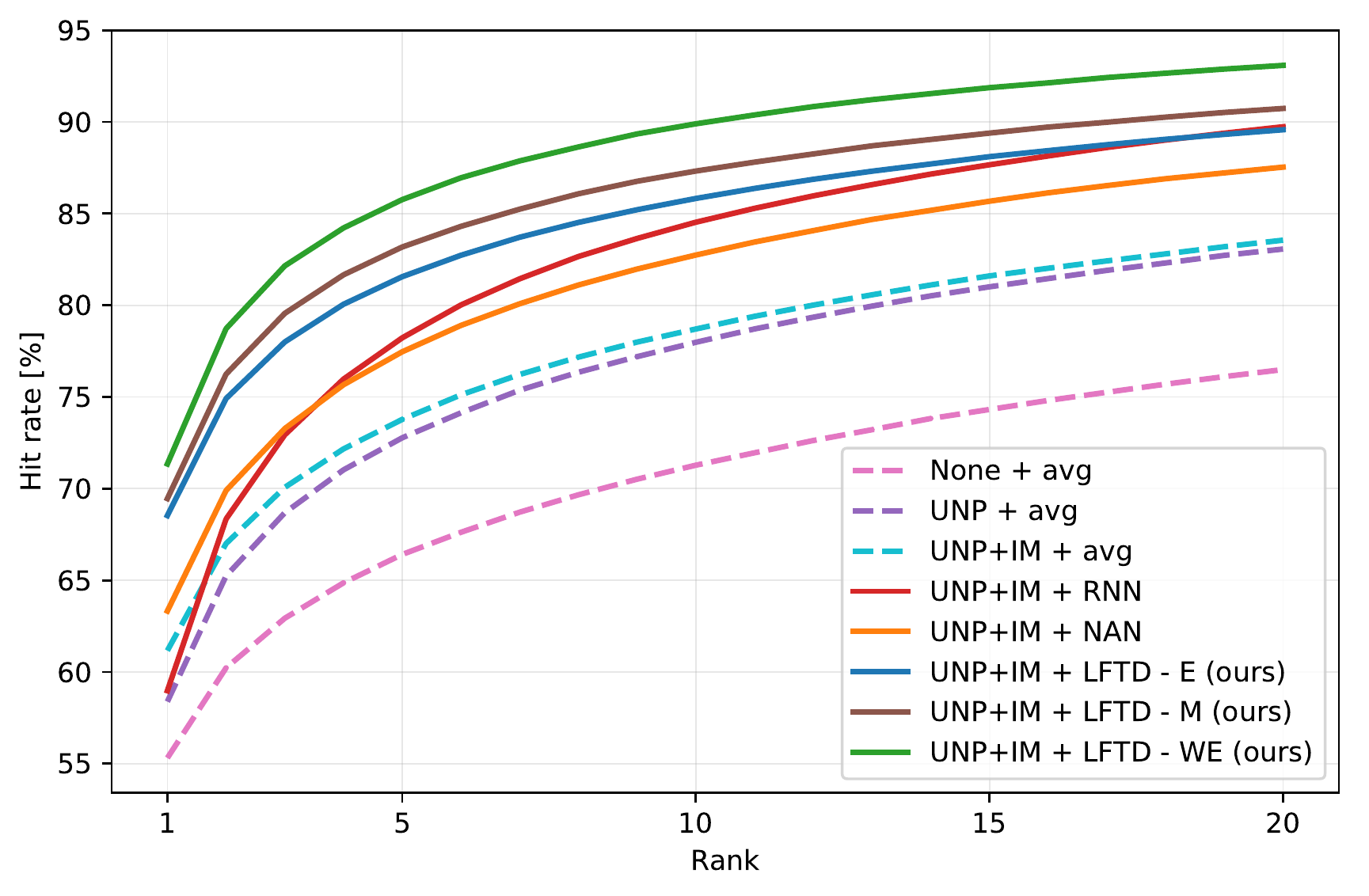}%
	\caption{Cumulative Matching Curve for different methods for vehicle re-identification on \dataset\ dataset. The methods use 128 dimensional feature vectors with the exception of avg which uses \avgFeatCars\ dimensional feature vectors. The methods use Euclidean distance with the exception of LFTD -- M (full Mahalanobis \citep{Shi2016}) and LFTD -- WE (Weighted Euclidean as proposed in Section \ref{sec:MethodologyMetrics}).} \label{fig:VehicleReIdCMC}
\end{figure}

%

The vehicle re-identification results can be found in Table~\ref{tab:VehicleReIdResults} and Cumulative Matching Curve (CMC) is shown in Figure~\ref{fig:VehicleReIdCMC}.
The results show several things.
First, both the Unpack (UNP) \citep{Sochor2016BoxCars} modification and image modifications (IM) \citep{Sochor2017BoxCars} improve the accuracy of vehicle re-identification. Second, all feature aggregation methods in the temporal domain (RNN \citep{McLaughlin2016}, NAN \citep{Yang2017}, LFTD) improve the accuracy when compared with the average pooling in the task of vehicle re-identification. Third, our method (LFTD) outperforms other methods for temporal aggregation (RNN \citep{McLaughlin2016}, NAN~\citep{Yang2017}). Finally, using other metrics than Euclidean also improves the accuracy. Our proposed Weighted Euclidean distance significantly outperforms the full Mahalanobis distance (as proposed by \cite{Shi2016}); and at the same time, our method has significantly lower time demands. It has time and memory complexity $\mathcal{O}(D)$ instead of $\mathcal{O}(D^2)$ for the full Mahalanobis distance, where $D$ is the dimensionality of the feature vectors.

\begin{figure}[t]
	\includegraphics[width=\linewidth]{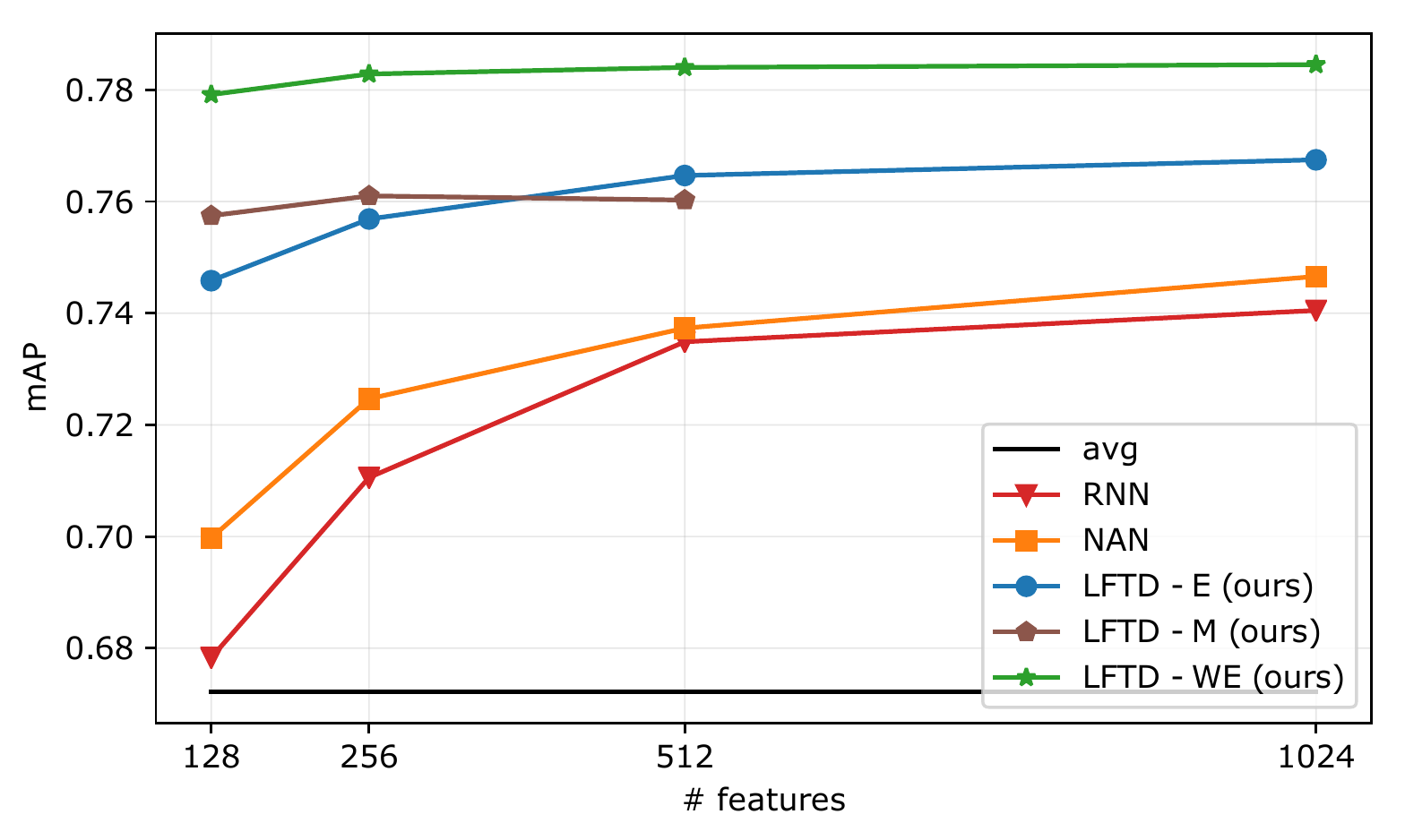}%
	\caption{Performance analysis of different methods for feature vector aggregation in temporal domain with changing number of features on \dataset\ dataset.  The avg pooling is shown only for visual comparison and uses \avgFeatCars\ dimensional feature vectors. We omitted LFTD -- M with 1024 features from evaluation because of long evaluation time (months) and performance drop of version with 512 features. All the methods use Euclidean distance with the exception of LFTD -- M (full Mahalanobis \citep{Shi2016}) and LFTD -- WE (Weighted Euclidean as proposed in Section~\ref{sec:MethodologyMetrics}).}
	\label{fig:VehicleReIdmAPAnalysis}
\end{figure}

\new{Our explanation of better performance of Weighted Euclidean distance instead of Mahalanobis distance is that there is not enough training data to train the full matrix M. This hypothesis is supported by Fig. \ref{fig:VehicleReIdmAPAnalysis} where the performance with Mahalanobis distance  does not increase and by the fact that 
	$\dfrac{\mathrm{tr}(|\mathbf{M}|)}{\sum{|\mathbf{M}|}}=0.997$
	, i.e. almost all the information in the matrix is on its diagonal.}

We were also curious how the accuracy changes with increasing the dimensionality of the feature vectors. As Figure~\ref{fig:VehicleReIdmAPAnalysis} shows, all methods improve with increasing dimensionality; however, the results are still similar. Our method LFTD with our proposed Weighted Euclidean distance is outperforming all other methods for all of the tested feature vector dimensionalities.

\subsection{Person Re-Identification}

To show that our method is applicable also outside the scope of vehicle re-identification, we evaluated it on the person re-identification task. We use two common datasets: iLIDS-VID \citep{Wang2014} and PRID-2011 \citep{Hirzer2011} as they are usually used by other methods for feature aggregation in temporal domain \citep{Yan2016,McLaughlin2016,Gao2016,Xu2017,Zhang2017,Chen2017,Zhou2017}.
Futrhermore, for fair comparison of proposed method, our work was also evaluated on the MARS dataset by \cite{Zheng2016mars}.

It should be noted that the subject of our study is the aggregation of features extracted on images by an existing feature extractor. That is why we include in the comparison those methods that do the same, not methods which use a significantly different method of image \textbf{feature extraction}.

\new{For the above reasons, we are not comparing our method with some of the currently published methods such as QAN \citep{Liu2017quality} or SpaAtn (DRSTA) \citep{Li2018diversity} as they are focusing on \emph{spatiotemporal} attention pooling, and because of that, they provide enhanced feature extraction.  In our work, we target fusion of existing feature extractors.  Besides that, their evaluation is not following the standard evaluation protocol used in the previous works and with the used datasets, as they are pre-training the networks on different types of image-based person re-identification tasks, thus their results are hardly comparable.}

\subsubsection*{iLIDS-VID and PRID-2011}

We always used a half of the dataset for training and the other half for testing. Therefore, the evaluation is done on 100 tracks (150 tracks) with PRID-2011 (iLIDS-VID) dataset. We used 10 random splits in the case of the PRID-2011 dataset, and the 10 published splits in the case of iLIDS-VID.

We used ResNet50 \citep{He2016} as the feature extractor from the images and trained it on the identification task by Adam optimizer with learning rate 0.0001 for 60 epochs with batch size 8, using standard augmentation techniques (random flip, rotation, and shift). We trained our method (LFTD) in a Siamese network by Adam optimizer with cross-validated learning rate for 150 epochs with batch size 8. We always used 16 time samples per track and contrastive loss margin 2. We also evaluate the average pooling with KISSME~\citep{Kostinger2012}  and XQDA~\citep{Liao2015} with cross-validated hyperparameters (regularization, and PCA reduction dimensionality in the case of KISSME).

\begin{table}[t]
	\centering
	\fontsize{6}{8}\selectfont
	\caption{Person re-identification results on PRID-2011 and iLIDS-VID dataset. The top-3 results are highlighted in the following way: \first{first}, \second{second}, and \third{third}. KISSME -- \cite{Kostinger2012}, XQDA -- \cite{Liao2015}. \new{LFTD metric used for this experiment is the standard Euclidean distance because of insufficient amount of training data for the Weighted Euclidean.}}
    \label{tab:PersonReIdResults}
	\begin{tabular}{l@{\hspace{2em}}rrrr@{\hspace{2em}}rrrr}
		\toprule
		\textbf{Method} & \multicolumn{4}{c}{\textbf{Hit@R (PRID)}} & \multicolumn{4}{c}{\textbf{Hit@R (iLIDS)}}\\
		&    1 &    5 &   10 &   20 &    1 &    5 &   10 &   20 \\
		\midrule
		\cite{Yan2016} & 58.2 & 85.8 & 93.4 & 97.9 & 49.3 & 76.8 & 85.3 & 90.0\\
		\cite{McLaughlin2016} & 70.0 & 90.0 & 95.0 & 97.0 & 58.0 & 84.0 & 91.0 & 96.0\\
		\cite{Gao2016} & 68.6 & \second{94.6} & \third{97.4} & 98.9 & 55.0 & \first{87.5} & \second{93.8} & \second{97.2}\\
		\cite{Xu2017} & 77.0 & \first{95.0} & \first{99.0} & \third{99.0} & 62.0 & \third{86.0} & \first{94.0} & \first{98.0} \\
		\cite{Zhang2017} & 72.8 & 92.0 & 95.1 & 97.6 & 55.3 & 85.0 & \third{91.7} & 95.1\\
		\cite{Chen2017} & 77.0 & 93.0 & 95.0 & 98.0 & 61.0 & 85.0 & \first{94.0} & \third{97.0}\\
		\cite{Zhou2017} & \third{79.4} & \third{94.4} & --- & \first{99.3} & 55.2 & \second{86.5} & --- & \third{97.0}\\
		\cite{Zhang2017learning} & 60.2 & 85.1 & --- & 94.2 & 83.3 & 93.3 & --- & 96.7 \\
		\midrule
		avg                  & 69.4 & 90.5 & 95.0 & 97.6 & 46.3 & 71.4 & 80.2 & 87.6 \\
		avg + KISSME             & 70.5 & 91.0 & 95.1 & 97.7 & 56.1 & 79.0 & 87.9 & 93.9 \\
		avg + XQDA                & 75.6 & 94.3 & \second{98.2} & \third{99.0} & 59.5 & 83.7 & 90.3 & 96.2 \\
		\midrule
		LFTD (128)  & 79.2 & 92.4 & 95.8 & 98.4 & 61.4 & 79.9 & 87.8 & 93.5 \\
		LFTD (256)  & \third{79.4} & 93.7 & 96.8 & 98.6 & \third{62.8} & 82.1 & 88.1 & 94.1 \\
		LFTD (512)  & \first{80.2} & \second{94.6} & 97.3 & 98.9 & \second{63.5} & 83.3 & 89.5 & 94.9 \\
		LFTD (1024) & \second{80.0} & 93.9 & \third{97.4} & \second{99.2} & \first{63.7} & 82.9 & 90.0 & 94.7 \\
		\bottomrule
	\end{tabular}
\end{table}

We used standard Euclidean distance as the metric for our algorithm, because the number of training data in the datasets is rather low and the accuracy did not improve further with other distances. This is caused mainly by insufficient amount of training data because the network was able to re-identify the training tracks without any error already with the standard Euclidean distance.

The results can be found in Table~\ref{tab:PersonReIdResults} and as the table shows, LFTD significantly increases the performance compared to average pooling or average pooling with other metric learning (KISSME, XQDA). The results also show that our method outperforms other methods for feature aggregation in temporal domain \citep{Yan2016,McLaughlin2016,Gao2016,Xu2017,Zhang2017,Chen2017,Zhou2017} in Hit@1.

\new{Evaluation of KISSME or XQDA metrics together with the features produced by the proposed LFTD method is not included in the results because of lacking relevancy of such comparison. LFTD produces features dependent on the metric used during training and it generates different feature vector representations for different metrics involved (Euclidean, Weighted Euclidean, Mahalanobis).}


\subsubsection*{MARS dataset}
Features published by the authors of the dataset were used in our experiments. The network was trained on the training part of the published features in the Siamese setting for 30 epochs with batch size 32 with Contrastive Loss and Adam optimizer. Hard negative mining was employed during the training, and all positive pairs and 20 hardest negative pairs were presented to the network in one epoch during the training. Values of learning rate and loss margin were fine-tuned for each variant individually. All variants of our method evaluated on the dataset can be found in Table \ref{tab:MARSReIdResults}. It should be noted that \emph{Baseline} is the variant (\emph{IDE, average pooling, Euclidean distance, single query}) reported by the authors \cite{Zheng2016mars}.

\begin{table}[t]
	\centering
	\scriptsize
	\caption{Person re-identification results on MARS dataset. Baseline is the variant (\emph{IDE, average pooling, Euclidean distance, single query}) reported by authors of the dataset \citep{Zheng2016mars}.
	\newline * - RNN-CNN \citep{McLaughlin2016} trained by \cite{Xu2017}.
	\SecondRev{- Experiments on Mars lack of recent baselines.}}
	\label{tab:MARSReIdResults}
	\begin{tabular}{lrrrrr}
		\toprule
		& & \multicolumn{4}{c}{Hit@Rank} \\
		Variant            	& mAP		& 1			& 5  	 	& 10		& 20   \\
		\midrule
		Baseline           	& 0.424		& 60.0		& 77.9	& - 		& 87.9 \\
RNN-CNN* \citep{Xu2017}  	& -     	& 40.0		& 64.0	& 70.0		& 77.0 \\
ASTPN \citep{Xu2017} 		& -			& 44.0		& 70.0	& 74.0		& 81.0 \\
\cite{Zhang2017learning}	& -			& 55.5		& 70.2	& -			& 80.2 \\
		\midrule
		LFTD - E (512)   	& 0.481		& 65.5  	& 80.3	& 85.5   	& 89.4 \\
		LFTD - E (1024)   	& 0.483		& 65.9   	& 80.7  & 84.8  	& 89.2 \\
		LFTD - WE (512) 	& 0.488   	& 66.1   	& 81.0  & 85.4   	& 89.8 \\
		LFTD - WE (1024)  	& 0.489   	& 66.4   	& 81.5  & 85.9   	& 89.8 \\
		\bottomrule
	\end{tabular}
\end{table}

%% file: 2018-CVIU-FeatureImportance-6-Conclusions.tex
\section{Conclusions}

We proposed a new scheme for extracting feature vectors for the whole tracks of multiple observations of an object (vehicle, person) of interest in the re-identification task.
Our method can work with arbitrary per-image features (e.g. feature vectors from ResNet50 or Inception-ResNet-v2). Based on such feature vectors we learn a considerably shorter (128 features) per-track feature vector by using the newly proposed LFTD (Learning Features in Temporal Domain).
We also propose to use a~different distance metric for comparing the feature vectors -- $\mathbf{WE}$ (Weighted Euclidean).  It is based on the Mahalanobis distance, whose learned matrix $\mathbf{M}$ is made diagonal.  This proposed distance metric is much cheaper in terms of computational and memory resources ($\mathcal{O}(D)$ instead of $\mathcal{O}(D^2)$ in the case of the full Mahalanobis metric), but at the same time, it is better at solving the re-identification task.

The results show that the increase of HIT@1 by using the LFTD was 7.3 percentage points for the vehicle re-identification task compared to average pooling, and 17.4 percentage points for the person re-identification with the iLIDS-VID dataset and up to 6.4 percentage points on the MARS dataset. The Weighted Euclidean metric further increased HIT@1 by other 2.8 percentage points in case of vehicle re-identification.

We collected and annotated a vehicle re-identification dataset \dataset\ for development and evaluation of vehicle re-identification systems and we make it public.  It contains \vehiclesAll\ unique vehicles, \tracksAll\ observed tracks, and \pairsAll\ positive pairs, taken from various angles -- not just from the front or rear.